\documentclass[10pt,twocolumn,letterpaper]{article}

\usepackage{3dv}
\usepackage{times}
\usepackage{graphicx}

\usepackage{caption}
\usepackage{comment}
\usepackage{amsmath} %
\usepackage{amssymb}
\usepackage{array}
\usepackage{color}
\usepackage{epsfig}
\usepackage{multirow}
\usepackage{subcaption}
\usepackage{float}
\usepackage{xspace}
\usepackage{lineno}
\usepackage[pagebackref=true,breaklinks=true,letterpaper=true,colorlinks,bookmarks=false]{hyperref}

\makeatletter
\def\blfootnote{\gdef\@thefnmark{}\@footnotetext}
\makeatother

\threedvfinalcopy

\newcommand{\bx}{\mathbf{x}}

\newcommand{\bS}{\mathbf{S}}
\newcommand{\bs}{\mathbf{s}}

\newcommand{\bt}{\mathbf{t}}

\newcommand{\bu}{\mathbf{u}}

\newcommand{\bD}{\mathbf{D}}

\newcommand{\bC}{\mathbf{C}}

\newcommand{\bc}{\mathbf{c}}

\newcommand{\bT}{\mathbf{T}}

\newcommand{\bd}{\mathbf{d}}
\newcommand{\br}{\mathbf{r}}

\newcommand{\nR}{\mathbb{R}}

\newcommand{\nS}{\mathbb{S}}

\newcommand{\cL}{\mathcal{L}}

\makeatletter
\DeclareRobustCommand\onedot{\futurelet\@let@token\@onedot}
\def\@onedot{\ifx\@let@token.\else.\null\fi\xspace}
\def\eg{e.g\onedot}

\def\etal{\textit{et~al}\onedot} 
\def\Fig{Fig\onedot}   
\makeatother

\newcommand{\figref}[1]{\Fig~\ref{#1}}
\newcommand{\secref}[1]{Section~\ref{#1}}

\renewcommand{\eqref}[1]{Eq.~\ref{#1}}
\newcommand{\tabref}[1]{Table~\ref{#1}}

\newcommand*\rot{\rotatebox{90}}
\newcommand{\boldparagraph}[1]{\vspace{0.2cm}\noindent{\bf #1:} }

\newif\ifcomment
\commenttrue
\ifcomment
	\newcommand{\ag}[1]{ \noindent {\color{red} {\bf Andreas:} {#1}} }
	\newcommand{\yl}[1]{ \noindent {\color{cyan} {\bf Yiyi:} {#1}} }

\else
	\newcommand{\ag}[1]{}
	\newcommand{\yl}[1]{}
\fi

\ifthreedvfinal\fi

\begin{document}
\title{Panoptic NeRF: 3D-to-2D Label Transfer for Panoptic Urban Scene Segmentation}

\author{Xiao Fu$^{1*}$\and
Shangzhan Zhang$^{1*}$ \and
Tianrun Chen$^{1}$ \and 
Yichong Lu$^{1}$  \and 
Lanyun Zhu$^{2}$ \and
Xiaowei Zhou$^{1}$  \hspace*{1cm}
Andreas Geiger$^{3}$ \hspace*{1cm} 
Yiyi Liao$^{1\dagger}$ \vspace{0.2cm}\\
$^{1}$Zhejiang University 
\quad 
$^{2}$Singapore University of Technology and Design 
\\
$^{3}$University of Tübingen and MPI-IS, Tübingen
}

\maketitle
\thispagestyle{empty}
\begin{figure*}[h]
\centering
\includegraphics[width=\linewidth]{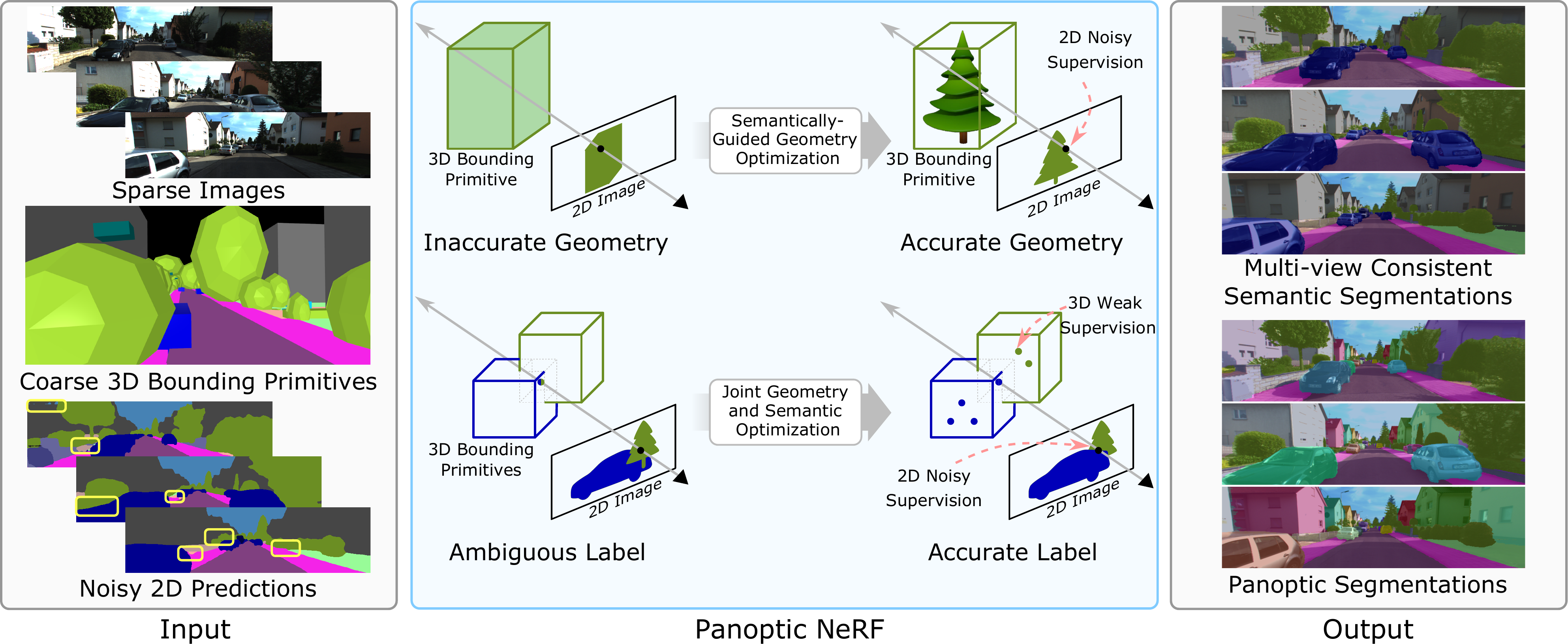}
\caption{\textbf{Panoptic NeRF} takes as input a set of sparse images, coarse 3D bounding primitives and noisy 2D predictions (yellow boxes highlight inaccurate predictions). By inferring in 3D space, it generates semantic and instance labels in the 2D image space via volume rendering. We propose semantically-guided geometry optimization to improve the underlying geometry. We further resolve label ambiguity at intersection regions of the 3D bounding primitives intersect via joint geometry and semantic optimization, enabling rendering consistent and accurate panoptic labels at multiple views.}
\label{fig:teaser}
\vspace{-0.3cm}
\end{figure*}

\blfootnote{$^*$Equal contribution. $^\dagger$Corresponding author.}
\blfootnote{Project Page: \href{https://fuxiao0719.github.io/projects/panopticnerf/}{https://fuxiao0719.github.io/projects/panopticnerf/}}

\begin{abstract}
Large-scale training data with high-quality annotations is critical for training semantic and instance segmentation models. Unfortunately, pixel-wise annotation is labor-intensive and costly, raising the demand for more efficient labeling strategies. In this work, we present a novel 3D-to-2D label transfer method, Panoptic NeRF, which aims for obtaining per-pixel 2D semantic and instance labels from easy-to-obtain coarse 3D bounding primitives. Our method utilizes NeRF as a differentiable tool to unify coarse 3D annotations and 2D semantic cues transferred from existing datasets. We demonstrate that this combination allows for improved geometry guided by semantic information, enabling rendering of accurate semantic maps across multiple views. Furthermore, this fusion process resolves label ambiguity of the coarse 3D annotations and filters noise in the 2D predictions. By inferring in 3D space and rendering to 2D labels, our 2D semantic and instance labels are multi-view consistent by design. Experimental results show that Panoptic NeRF outperforms existing label transfer methods in terms of accuracy and multi-view consistency on challenging urban scenes of the KITTI-360 dataset. 

\end{abstract}
\section{Introduction}
Semantic instance segmentation is an important perception task for autonomous driving. It is widely acknowledged that large-scale training data with high-quality annotations is critical to propel the performance of segmentation models. However, manual annotation of pixel-accurate segmentation masks is highly expensive and time-consuming. For example, annotating all instances in a single street scene image requires up to 1.5 hours~\cite{liao2021kitti}. 
Recently, a few urban datasets propose to annotate in 3D space using coarse bounding primitives (e.g., cuboids and ellipsoids) and transfer 3D labels to 2D~\cite{xie2016semantic,liao2021kitti,huang2019apolloscape}, significantly reducing the annotation time to 0.75 minutes per image~\cite{liao2021kitti}.  Besides, it is often easier to separate instances in 3D rather than in 2D image space (e.g., pedestrian in front of building). Thus, there is an increasing demand for accurately transferring coarse 3D annotations to 2D semantic and instance labels.

There are only a few existing attempts in this direction. Huang \etal~\cite{huang2019apolloscape} perform manual post-processing to obtain accurate 2D labels based on 3D coarse annotations. Another line of work additionally leverages 2D image cues (\eg, noisy 2D semantic predictions) to avoid human intervention, achieved by combining 3D annotations and 2D image cues based on conditional random fields (CRF)~\cite{xie2016semantic,liao2021kitti}. This CRF-based approach relies on intermediate 3D reconstructions for projecting non-occluded 3D points to 2D, and then performs inference in \textit{2D image space}. The 3D reconstruction cannot be jointly optimized in the CRF model and thus erroneous reconstruction leads to inaccurate label transfer results. To alleviate this problem, we propose Panoptic NeRF, a novel label transfer method built on NeRF~\cite{mildenhall2020nerf} that infers geometry and semantic jointly in \textit{3D space} to render dense 2D semantic and instance labels, i.e., panoptic segmentation labels~\cite{kirillov2019panoptic} (see \figref{fig:teaser}).

A na\"ive solution is to first train a vanilla NeRF model, and then render segmentation maps based on semantic/instance labels determined by the 3D bounding primitives. However, as illustrated in \figref{fig:teaser}, inaccurate geometric reconstruction leads to wrong semantic/instance maps, yet it is hard to obtain accurate geometry using the vanilla NeRF in the driving scenario where input views are sparse. Furthermore, label ambiguity at overlapping regions of the 3D bounding primitives also yields inaccurate 2D labels. 

In this work, we aim to tackle both challenges. Inspired by~\cite{xie2016semantic,liao2021kitti}, we combine 3D annotations and noisy 2D semantic predictions transferred from existing datasets to fully automate the label transfer process.
Specifically, our method consists of a radiance field and \textit{dual semantic fields}, supervised by 3D and 2D weak semantic information as well as posed 2D RGB images.
To improve the underlying geometry, we propose a semantically-guided geometry optimization strategy based on a \textit{fixed semantic field} determined by the 3D bounding primitives. 
With the semantic field fixed, we demonstrate that the geometry can be improved guided by noisy 2D semantic predictions.
The semantic rendering is then further refined by joint geometry and semantic optimization, where a \textit{learned semantic field} is adopted to fuse information of the  3D bounding primitives and the 2D noisy predictions. 
As evidenced by our experiments, this fusion procedure is able to resolve the label ambiguity of the 3D bounding primitives and largely eliminate noise in the 2D predictions. Furthermore, Panoptic NeRF enables rendering globally consistent 2D instance maps across multiple frames, where each object has a unique instance index determined by the 3D bounding primitives. Utilizing 3D bounding primitives of the recently released KITTI-360 dataset, Panoptic NeRF outperforms existing 3D-to-2D and 2D-to-2D label transfer methods, providing a promising approach to efficiently develop large-scale and densely labeled datasets for autonomous driving.

We summarize our contributions as follows: 
    1) We propose to perform 3D-to-2D label transfer by inferring in the 3D space. This allows us to unify easy-to-obtain 3D bounding primitives and noisy 2D semantic predictions in a single model, yielding high-quality panoptic labels.  
    2) By leveraging a novel dual formulation of the semantic fields, Panoptic NeRF effectively improves the geometric reconstruction given sparse views, yielding accurate object boundaries. Moreover, it is able to resolve label ambiguities and eliminates label noise based on the improved geometry.
    3) Our Panoptic NeRF achieves superior performance compared to existing label transfer methods in terms of both semantic and instance predictions. 
    Furthermore, our 2D semantic and instance labels are multi-view and spatio-temporally consistent by design. Finally, our method enables rendering RGB images and semantic/instance labels at novel viewpoints.

\section{Related Work}
\boldparagraph{Urban Scene Segmentation} Semantic instance segmentation is a critical task for autonomous vehicles~\cite{ohn2020learning,Zhou2019SR}. 
Learning-based algorithms have achieved compelling performance~\cite{chen2017deeplab,li2020semantic,zhao2017pyramid}, but rely on large-scale training data. Unfortunately, annotating images at pixel level is extremely time-consuming and labor-intensive, especially for instance-level annotation. While most urban datasets provides labels in 2D image space~\cite{brostow2009semantic,cordts2016cityscapes,neuhold2017mapillary,Kim2020CVPR,weber2021step}, autonomous vehicles are usually equipped with 3D sensors~\cite{geiger2012we,caesar2020nuscenes,Geyer2020ARXIV,huang2019apolloscape,liao2021kitti}. KITTI-360~\cite{liao2021kitti} demonstrates that annotating the scene in 3D can significantly reduce the annotation time. However, transferring coarse 3D labels to 2D remains challenging. In this work, we focus on developing a novel 3D-to-2D label transfer method, exploiting recent advances in neural scene representations.

\boldparagraph{Label Transfer} There have been several attempts at improving label efficiency for individual frames~\cite{liu2011nonparametric,Guillaumin2014IJCV,castrejon2017annotating,acuna2018efficient,ling2019fast,andriluka2018fluid}. In this paper we focus on efficient labeling of video sequences. Existing works in this area can be divided into two categories: 2D-to-2D and 3D-to-2D. 
2D-to-2D label transfer approaches reduce the workload by propagating labels across 2D images~\cite{pathak2015constrained,hong2016learning,papandreou2015weakly,pinheiro2015image,ganeshan2021warp}, whereas 3D-to-2D methods exploit additional information in 3D for efficient labeling~\cite{huang2019apolloscape,martinovic20153d,mustafa2017semantically,xiao2009multiple,bruls2018mark}. To obtain dense labels in 2D image space, some works~\cite{liao2021kitti,xie2016semantic} perform per-frame inference jointly over the 3D point clouds and 2D pixels using a non-local multi-field CRF model. However, these methods require reconstructing a 3D mesh to project 3D point clouds to 2D. As it is treated as a pre-processing step, the mesh reconstruction is not jointly optimized in the CRF model. Thus, inaccurate reconstruction hinders label transfer performance. In contrast, Panoptic NeRF provides a novel end-to-end method for 3D-to-2D label transfer where geometry and semantic estimations are jointly optimized.

\boldparagraph{Coordinate-based Neural Representations} Recently, coordinate-based neural representations has received wide attention.
in many areas, including 3D reconstruction~\cite{park2019deepsdf,jiang2020local,genova2020local,mescheder2019occupancy,niemeyer2020differentiable,yariv2020multiview,gropp2020implicit,oechsle2021unisurf,peng2020convolutional,sitzmann2020implicit}, novel view synthesis~\cite{mildenhall2020nerf,barron2021mip,kellnhofer2021neural,martin2021nerf}, and 3D generative modeling~\cite{gu2021stylenerf,meng2021gnerf,schwarz2020graf}. In this paper, we focus on utilizing coordinate-based representations to estimate the semantics of the scene. Towards this goal, NeSF~\cite{vora2021nesf} focuses on generalizable semantic field learning from density grids supervised by 2D GT labels, but we concentrate on the 3D-to-2D label transfer task without access to 2D GT. A closely related work Semantic NeRF~\cite{zhi2021place} explores NeRF for semantic fusion. However, Semantic NeRF takes as input ground truth 2D labels or synthetic noise labels, which struggles to produce correct labels given real-world predictions from pre-trained 2D models. Moreover, Semantic NeRF operates in indoor scenes with dense RGB inputs and degenerates in challenging outdoor driving scenarios with sparse input views. Finally, Semantic NeRF is limited to rendering semantic labels, whereas our method can render panoptic labels. A concurrent work PNF~\cite{KunduCVPR2022PNF} allows for rendering papotic labels. While PNF focuses on parsing the scene using known classes of the Cityscapes dataset~\cite{cordts2016cityscapes} based on pre-trained segmentation and detection models, we aim to transfer 3D annotations to 2D image space for arbitrary classes to enable the development of new datasets, \eg, providing instance labels for buildings that are not available in Cityscapes.

\section{Background}
\boldparagraph{NeRF}
NeRF~\cite{mildenhall2020nerf} models a 3D scene as a continuous neural radiance field $f_{\theta}$. Specifically, it maps a 3D coordinate $\bx$ and a viewing direction $\bd$ to a volume density $\sigma$ and an RGB color value $\bc$:
\begin{equation}
f_{\theta}: (\bx\in\nR^3, \bd\in\nS^2) \mapsto (\sigma\in\nR^+,\bc\in\nR^3)
\label{eq:nerf}
\end{equation}

Let $\mathbf{r}(t)=\mathbf{o}+t \mathbf{d}$ denote a camera ray. The color at the corresponding pixel can be obtained by volume rendering
\begin{equation}
\bC(\br)=\sum_{i=1}^{N} T_i (1-\exp (-\sigma_i\delta_i)) \bc_i \ , \  T_i = \exp \left(-\sum_{j=1}^{i-1} \sigma_j \delta_j \right)
\end{equation}
where $\sigma_i$ and $\bc_i$ denote the density and color value at a point $i$ sampled along the ray,  $T_i$ denotes the transmittance at the sample point, and $\delta_{k}=t_{i+1}-t_{i}$ is the distance between adjacent samples. Let $\pi$ denote the volume rendering process of one ray. Enabled by volume rendering, NeRF learns $f_\theta$ from a set of 2D RGB images with known camera poses.

\boldparagraph{Problem Formulation}\label{sec:formulation}
As shown in \figref{fig:teaser}, Panoptic NeRF aims to transfer coarse 3D bounding primitives to dense 2D semantic and instance labels. In addition to a sparse set of posed RGB images, we assume a set of 3D bounding primitives $\beta=\left\{B_{k}\right\}_{k=1}^{K}$ to be available. These 3D bounding primitives cover the full scene in the form of cuboids, ellipsoids and extruded polygons. Each 3D bounding primitive $B_{k}$ has a \textit{semantic} label, belonging to either ``stuff'' or ``thing''. For ``thing'' classes, $B_{k}$ is additionally associated with a unique \textit{instance} ID.
We further apply a pre-trained semantic segmentation model to the RGB images to obtain a 2D semantic prediction for each image.
With this input information, our primary goal is to generate multi-view consistent, semantic and instance labels at the input frames, whereas rendering RGB images and panoptic labels at novel viewpoints is also enabled by inferring in the 3D space.
\section{Methodology}

\begin{figure*}[tb]
\centerline{\includegraphics[width=\textwidth]{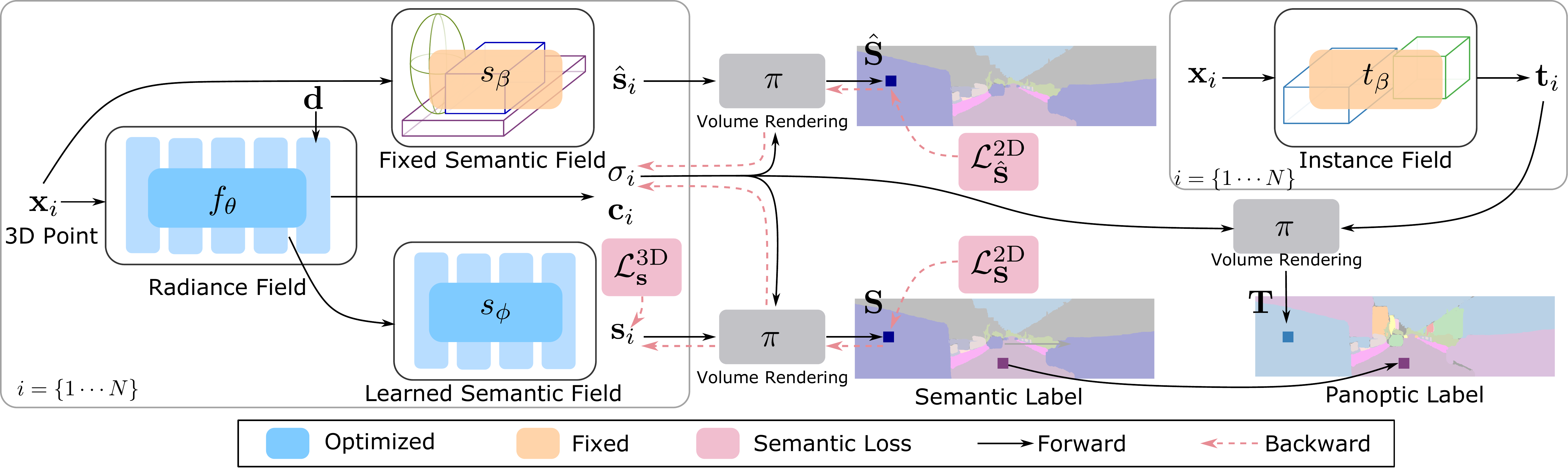}}
\vspace{-0.2cm}
\caption{\textbf{Method Overview}. \textit{Left} (Semantic Segmentation):  At each 3D location $\bx_i$, $i = \{1\cdots N\}$, sampled along a ray, we leverage dual semantic fields to obtain two semantic categorical distributions, $\hat{\bs}_i$ and ${\bs}_i$. The 3D semantic distributions are accumulated along the ray and projected to 2D image space via volume rendering, resulting in $\hat{\bS}$ and ${\bS}$. The semantic losses applied to both semantic fields improve 1) the volume density $\sigma_i$ and 2) the 3D semantic predictions ${\bs}_i$. \textit{Right} (Panoptic Segmentation):  Our method allows for rendering panoptic labels by combining the learned semantic field and a fixed instance field determined by the 3D bounding primitives.} 
\vspace{-0.2cm}
\label{fig:pipeline}
\end{figure*}

Panoptic NeRF provides a novel method for label transfer from 3D to 2D. 
\figref{fig:pipeline} gives an overview of our method. We first map a 3D point $\bx$ to a density $\sigma$ and a color value $\bc$ using a radiance field, as well as two semantic categorical distributions $\hat{\bs}$ and $\bs$ based on our dual semantic fields (\secref{sec:dual_semantic}). Correspondingly, for each camera ray, two semantic categorical distributions $\hat{\bS}$ and $\bS$ in 2D image space via volume rendering $\pi$ are obtained. Based on semantic losses in both 3D and 2D space (\secref{sec:loss}), the fixed semantic field $s_{\beta}$ serves to improve geometry, while the learned semantic field $s_{\phi}$ results in improved semantics.
With the 3D bounding primitives, we further define a fixed instance field $t_\beta$ that allows for rendering panoptic label $\bT$ when combined with the learned semantic field (\secref{sec:panoptic_rendering}).

\subsection{Dual Semantic Fields} \label{sec:dual_semantic}
To jointly improve geometry and  semantics, we define dual semantic fields, one is determined by the 3D bounding primitives $\beta$ and the other is learned by a semantic head $\phi$
\begin{equation}
s_{\beta}: \bx\in\nR^3 \mapsto \hat{\bs}\in\nR^{M_s} \quad~~ s_{\phi}: \bx\in\nR^3  \mapsto{\bs}\in\nR^{M_s}
\end{equation}
where $M_s$ denotes the number of semantic classes.  In combination with the volume density of the radiance field $f_\theta$, two semantic distributions $\hat{\bS}({\br})$ and  ${\bS}({\br})$ can be obtained at each camera ray $\br$ via the volume rendering operation $\pi$: %

\vspace{-0.2cm}
\begin{equation}
\begin{aligned}
\hat{\bS}(\br)=\sum_{i=1}^{N} T_i (1-\exp (-\sigma_i\delta_i))  \hat{\bs}_i \\
{\bS}(\br)=\sum_{i=1}^{N} T_i (1-\exp (-\sigma_i\delta_i))  {\bs}_i
\end{aligned}
\label{eq:semantic_rendering}
\end{equation}

Note that $\hat{\bS}(\br)$ and ${\bS}(\br)$ are both normalized distributions when $\sum_{i=1}^{N} T_i (1-\exp (-\sigma_i\delta_i))=1$. We set the background class to sky if $\sum_{i=1}^{N} T_i (1-\exp (-\sigma_i\delta_i))<1$. We apply losses to both $\hat{\bS}({\br})$ and ${\bS}({\br})$ for training.
During inference, the semantic label is determined as the class of the maximum probability in ${\bS}(\br)$. 

\boldparagraph{Fixed Semantic Field}
If $\bx$ is uniquely enclosed by a 3D bounding primitive $B_k$, $\hat{\bs}$ is a fixed one-hot categorical distribution of the category of $B_k$. For a point $\bx$ enclosed by multiple 3D bounding boxes of different semantic categories, we assign equal probability to these plausible categories and $0$ to the others. As explained in \secref{sec:loss}, the semantic field $s_{\beta}$ is able to improve the geometry but cannot resolve the label ambiguity at the overlapping region. 

\boldparagraph{Learned Semantic Field}
We add a semantic head parameterized by $\phi$ to NeRF to learn the semantic distribution ${\bs}$. We apply a softmax operation at each 3D point to ensure that ${\bs}$ is a categorical distribution. The detailed network structure can be found in the supplementary material.

\subsection{Loss Functions} \label{sec:loss}

\begin{figure}[tb]
\centerline{\includegraphics[width=.48\textwidth]{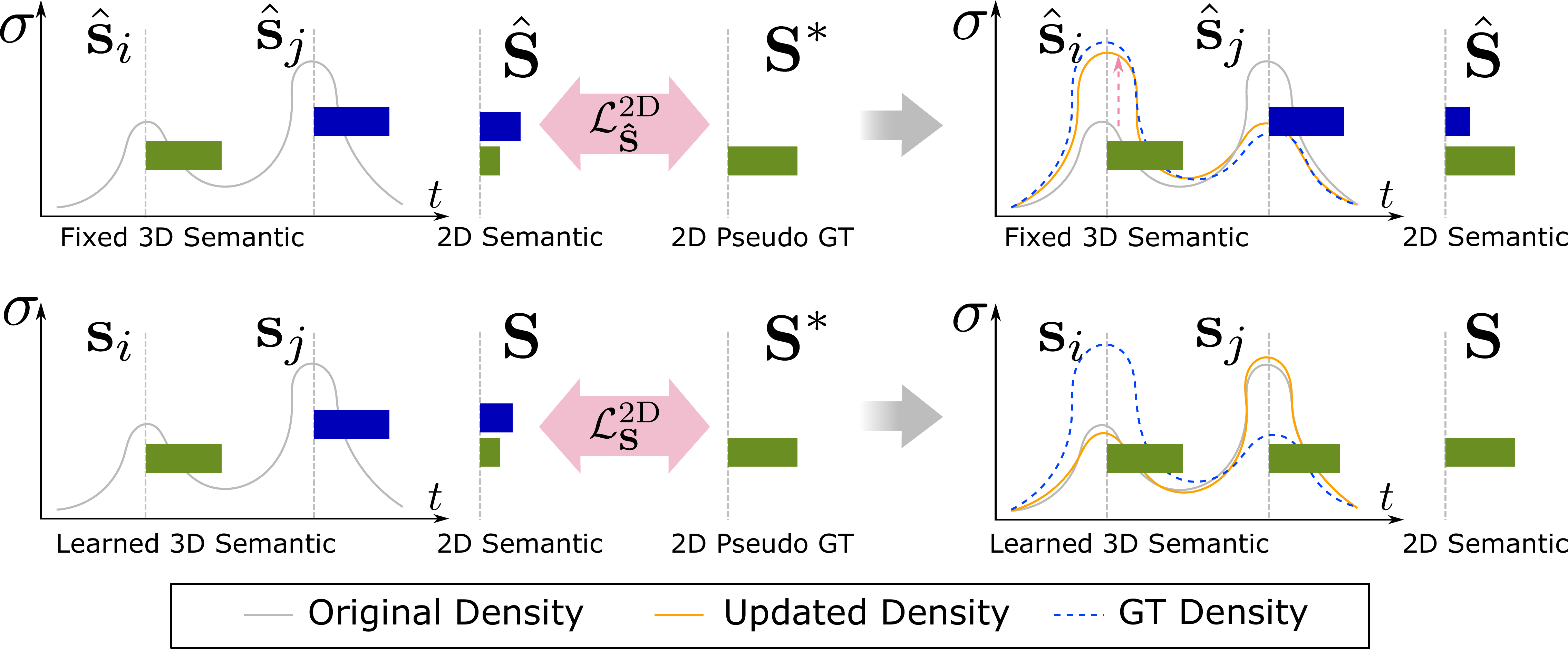}}
\caption{\textbf{Semantically-Guided Geometry Optimization}. The top row illustrates a single ray of the fixed semantic field $s_\beta$, where $\cL^{\text{2D}}_{\hat{\bS}}$ can only update the underlying geometry as the semantic distribution  $\hat{\bs}$ is fixed. 
The second row shows a single ray of the learned semantic field $s_\phi$. In this case, the network can ``cheat'' by adjusting the semantic prediction ${\bs}$ to satisfy $\cL^{\text{2D}}_{\bS}$ instead of updating the density $\sigma$. }
\label{fig:loss}
\end{figure}

\boldparagraph{Semantically-Guided Geometry Optimization}
In the driving scenario considered in our setting, the RGB images are sparse and the depth range is infinite.
We observe that the vanilla NeRF fails to recover reliable geometry in this setting.
However, we find that leveraging noisy 2D semantic predictions as pseudo ground truth is able to boost density prediction when applied to the fixed semantic fields $s_\beta$
\begin{equation}
\cL^{\text{2D}}_{\hat{\bS}}(\theta)=-\sum_{\mathbf{r} \in \mathcal{R}}\sum_{k=1}^{M_{s}} \bS^*_{k}(\mathbf{r}) \log \hat{\bS}_{k}(\mathbf{r})
\label{eq:loss_baseline1}
\end{equation}
where $\hat{\bS}_{k}(\br)$ denotes the probability of the camera ray $\br$ belonging to the class $k$, and $\bS^*_k(\br)$ denotes the corresponding pseudo-2D ground truth. As illustrated in \figref{fig:loss}, the key to improve density is to directly apply the semantic loss to the fixed semantic field $s_{\beta}$,
where $\cL^{\text{2D}}_{\hat{\bS}}$ can only be minimized by updating the density $\sigma$. 
\figref{fig:loss} shows that a correct $\bS^*$ increases the volume density of 3D points inside the correct bounding primitive and suppresses the density of others. When $\bS^*$ is wrong, the negative impact can be mitigated:
1) If $\bS^*$ does not match any bounding primitive along the ray, it has no impact on the radiance field $f_{\theta}$. 2) If $\bS^*$ exists in one of the bounding primitives along the ray, it means $\bS^*$ corresponds to an occluding/occluded bounding primitive with wrong depth. 
To compensate, we introduce a weak depth supervision $\cL_{d}$ based on stereo matching to alleviate the misguidance of $\cL^{\text{2D}}_{\hat{\bS}}$. Although  $\cL_{d}$ improves the overall geometry as shown in our ablation study, it fails to produce accurate object boundaries when used alone (see supplementary). Adding our semantically-guided geometry optimization yields more accurate density estimation as
pre-trained segmentation models usually perform well on frequently occurring classes, \eg, cars and roads. 

\boldparagraph{Joint Geometry and Semantic Optimization}
While enabling improved geometry, the 3D label of the overlapping regions remains ambiguous in the fixed semantic field. We leverage $s_{\phi}$ to address this problem by jointly learning the semantic and the radiance fields. We apply a cross-entropy loss $\cL^{\text{2D}}_{\bS}$ to each camera ray based on the filtered 2D pseudo ground truth, where $\bu(\mathbf{r})$ is set to 1 if $\bS^*(\mathbf{r})$ matches the semantic class of any bounding primitive along the ray and otherwise $0$. To further suppress noise in the 2D predictions, we add a per-point semantic loss $\cL^{\text{3D}}_{\bs}$ based on the 3D bounding primitives
\vspace{-0.2cm}
\begin{equation}
\begin{aligned}
\cL^{\text{2D}}_{\bS}(\theta,\phi) &=-\sum_{\mathbf{r} \in \mathcal{R}}\bu(\mathbf{r}) \sum_{k=1}^{M_{s}} \bS^*_{k}(\mathbf{r}) \log {\bS}_{k}(\mathbf{r}) \\
\cL^{\text{3D}}_{\bs}(\phi) &= - \sum_{\mathbf{r} \in \mathcal{R}}\sum_{i=1}^{N} \bu_i \sum_{k=1}^{M_s} \hat{\bs}_i^k \log {\bs}_i^k
\end{aligned}
\end{equation}
where $\bu_{i}$ is a per-point binary mask. $\bu_i$ is set to $1$ if (1) $\bx_i$ has a unique 3D semantic label and (2) the density $\sigma$ is above a threshold $\sigma_{th}$ to focus on the object surface.
As illustrated in \figref{fig:loss}, $\cL^{\text{2D}}_{\bS}(\theta,\phi)$ does not necessarily improve the underlying geometry as the network can simply adjust the semantic head $s_\phi$ to satisfy the loss. This behavior is also observed in novel view synthesis where NeRF does not necessarily recover good geometry when optimized for image reconstruction alone, specifically given sparse input views~\cite{Deng2021ARXIV,niemeyer2021regnerf}.

\boldparagraph{Total Loss}
Together, the total loss takes the form as
\begin{equation}
\cL =  \lambda_{\hat{\bS}} \cL^{\text{2D}}_{\hat{\bS}}  + \lambda_{{\bS}}\cL^{\text{2D}}_{\bS} + \lambda_{{\bs}}\cL^{\text{3D}}_{\bs}  + \lambda_{\bC}\cL_{p} + \lambda_{d}\cL_{d}
\end{equation}
where $\cL_{p}=\sum_{\mathbf{r} \in \mathcal{R}}\left\|\bC^*(\br)-\bC(\br)\right\|_{2}^{2}$ and $\cL_{d}=\sum_{\mathbf{r} \in \mathcal{R}}\left\|\bD^*(\br)-\bD(\br)\right\|_{2}^{2}$ denote the photometric loss and the depth loss, respectively. $\lambda_{\hat{\bS}}$, $\lambda_{{\bS}}$, $\lambda_{{\bs}}$, $\lambda_{\bC}$, and $\lambda_{d}$ are constant weighting parameters.
$\bC^*(\br)$ and $\bC(\br)$ are the ground truth and rendered RGB colors for ray $\br$. $\bD^*(\br)$ and $\bD(\br)$ are pseudo ground truth depth generated by stereo matching and rendered depth, respectively. Please refer to the supplementary for more details of $\bD^*$.

\subsection{Rendering of Panoptic Labels}\label{sec:panoptic_rendering}
Based on our learned semantic field $s_{\phi}$ and the 3D bounding primitives $\beta$ with instance IDs, we can easily render a panoptic segmentation map. 
Specifically, for a camera ray $\br$, the panoptic label directly takes the class with maximum probability in ${\bS}(\br)$ if it is a ``stuff'' class.
For ``thing'' classes, we render an instance distribution $\bT(\br)$ based on the bounding primitives $\beta$ to replace ${\bS}$ with $\bT$.
Our instance field is defined as follow
\begin{equation}
t_{\beta}: \bx\in\nR^3 \mapsto \bt\in\nR^{M_t}
\label{instance field}
\end{equation}
where ${M_t}$ is the number of the things in the scene and $\bt$ denotes a categorical distribution indicating which thing it belongs to. Note that $\bt$ is determined by the bounding primitives and is a one-hot vector if $\bx$ is uniquely enclosed by a bounding primitive of a thing. As overlap often occurs at the intersection of stuff and thing region, the bounding primitives of things rarely overlap with each other. Thus, this deterministic instance field leads to reliable performance in practice. 
To ensure that the instance label of this ray is consistent with the semantic class defined by ${\bS}$, we mask out instances belonging to other semantic classes by setting their probabilities to $0$ in $\bT$.

\subsection{Implementation Details} 

\boldparagraph{Sampling Strategy and Sky Modeling} \label{samplingstrategy} With the 3D bounding primitives covering the full scene, we sample points inside the bounding primitives to skip empty space. For each ray, we optionally sample a set of points to model the sky after the furthest bounding primitive.
More details regarding the sampling strategy can be found in the supplementary.
Our sampling strategy allows the network to focus on the non-empty region. As evidenced by our experiments, this is particularly beneficial in unbounded outdoor environments.  

\boldparagraph{Training} We optimize one Panoptic NeRF model per scene, using a single NVIDIA 3090. For each scene, we set the origin to the center of the scene. 
We use Adam~\cite{kingma2014adam} with a learning rate of 5e-4 to train our models. We set loss weights to $\lambda_{\hat{\bS}}=1, \lambda_{{\bS}}=1, \lambda_{{\bs}}=1, \lambda_{\bC}=1, \lambda_{d}=0.1$, and the density threshold to $\sigma_{th}=0.1$. We optimize the total loss $\cL$ for 80,000 iterations.

\begin{figure*}[!t]
 \centering
 \newcommand{\mywidth}{.95\textwidth}
 \setlength\tabcolsep{0.05em}
 \newcolumntype{P}[1]{>{\centering\arraybackslash}m{#1}}
 \def\arraystretch{0.50}
  \begin{tabular}{P{0.5em}P{0.5em}P{\mywidth}}
    \rot{\scriptsize{Input}}&& \includegraphics[width=\mywidth]{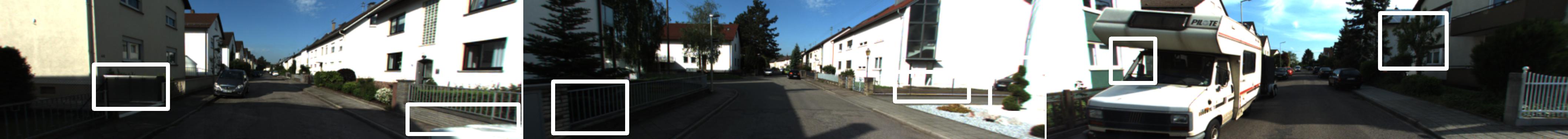} \\
     \rot{\tiny{S-NeRF+PSPNet*}}&\rot{\tiny{~\cite{zhi2021place}}}& \includegraphics[width=\mywidth]{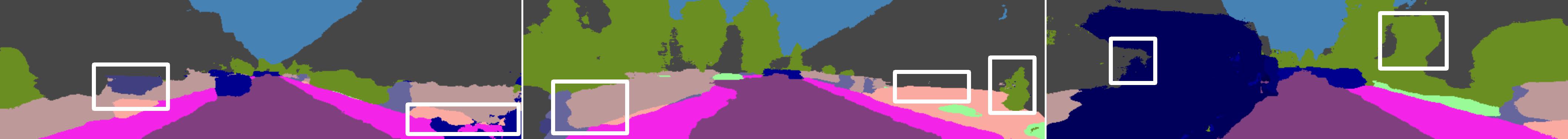}\\
     \rot{\tiny{PSPNet*}}&\rot{\tiny{~\cite{zhao2017pyramid}}}& \includegraphics[width=\mywidth]{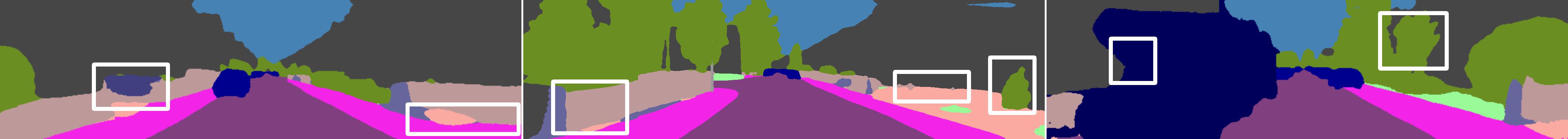}\\
     \rot{\tiny{3D-2D CRF}}&\rot{\tiny{~\cite{liao2021kitti}}}& \includegraphics[width=\mywidth]{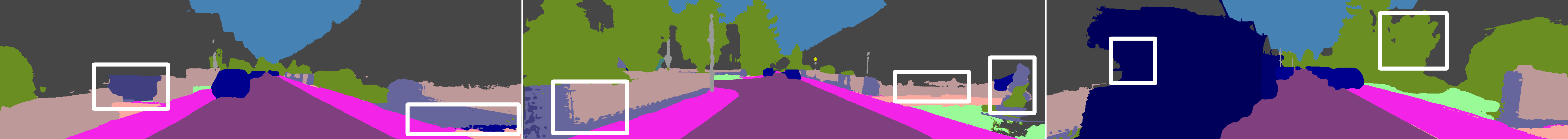} \\
    \rot{\scriptsize{Ours}}&& \includegraphics[width=\mywidth]{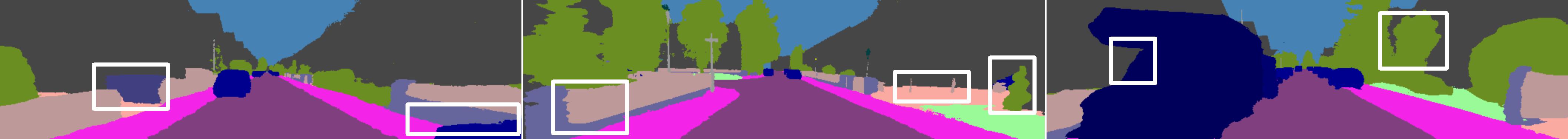}\\
   \rot{\scriptsize{GT}}&& \includegraphics[width=\mywidth]{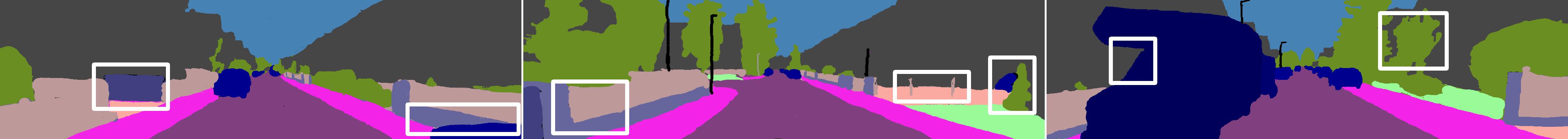}\\
 \end{tabular}
 \vspace{-0.2cm}
 \caption{
  \textbf{Qualitative Comparison of Semantic Label Transfer.}  Our method achieves superior performance in challenging regions compared to the baselines, e.g. in under- or over-exposed regions, by recovering the underlying 3D geometry, see box next to the building  (left) and building above the caravan (right).
 }
  \vspace{-0.3cm}
 \label{fig:semanitc}
\end{figure*}
 
\begin{table*}[t]
\centering
\resizebox{\textwidth}{!} {
\begin{tabular}{|c|cccccccccccccc|c|c|c|} \hline  
Method &Road &Park &Sdwlk &Terr &Bldg &Vegt &Car &Trler &Crvn &Gate &Wall &Fence &Box &Sky &mIoU &Acc &MC \\  \hline
FC CRF + Manual GT~\cite{krahenbuhl2011efficient} & 90.3 & 49.9 & 67.7 & 62.5 & 88.3 & 79.2 & 85.6 & 48.9 & 78.1 & 23.4 & 35.3 & 46.5 & 42.0 & 92.7 & 63.6 & 89.1 & 85.41 \\ 
S-NeRF + Manual GT~\cite{zhi2021place} & 87.0 & 35.8 & 64.7 & 58.2 & 83.4 & 76.3 & 70.3 & 93.5 & 76.5 & 41.4 & 44.0 & 52.6 & 29.0 & 92.0 & 64.6 & 86.8 & 88.98 \\ 
S-NeRF + PSPNet*~\cite{zhi2021place} & 94.6 & 52.9 & 77.7 & 65.0 & 88.0 & 80.8 & 87.9 & 58.3 & 86.0 & 36.0 & 44.1 & 56.8 & 42.2 & 90.9 & 68.6 & 90.5 & 93.42  \\ 
\hline
PSPNet*~\cite{zhao2017pyramid} & 95.5 & 49.7 & 77.5 & 66.7 & 88.9 & 82.4 & 91.6 & 46.5 & 83.1 & 24.2 & 43.3 & 51.3 & 51.1 & 89.3 & 67.2 & 90.7 & 91.79 \\ 
3D Primitives + GC & 81.7 & 31.0 & 45.6 & 22.5 & 59.6 & 56.7 & 63.0 & 61.7 & 37.3 & 61.6 & 28.8 & 50.6 & 39.5 & 50.3 & 49.3 & 73.4 & 86.56 \\ 
3D Mesh + GC & 91.7 & 53.1 & 67.2 & 31.4 & 81.3 & 72.1 & 85.2 & 93.5 & 86.0 & 65.2 & 40.7 & 59.7 & 54.4 & 65.6 & 67.7 & 86.0 & 94.99  \\ 
3D Point + GC & 93.5 & 59.0 & 76.1 & 37.2 & 82.0 & 74.1 & 87.5 & 94.7 & 85.7 & 66.7 & 59.4 & 65.9 & 58.6 & 68.0 & 72.0 & 87.9 & \textbf{96.51} \\ 
3D-2D CRF~\cite{liao2021kitti} & 95.2 & 64.2 & 83.8 & 67.9 & 90.3 & 84.2 & 92.2 & 93.4 & 90.8 & 68.2 & \textbf{64.5} & 70.0 & 55.8 & \textbf{92.8} & 79.5 & 92.8& 94.98 \\
\hline
Ours & \textbf{95.6} & \textbf{68.4} & \textbf{84.1} & \textbf{69.5} & \textbf{91.0} & \textbf{84.4} & \textbf{93.0} & \textbf{94.8} & \textbf{93.7} & \textbf{71.6} & 63.1 & \textbf{74.4} & \textbf{59.7} & 91.7 & \textbf{81.1} & \textbf{93.2} & 95.02\\ \hline 
\end{tabular}
}
\vspace{-0.2cm}
\caption{\textbf{Quantitative Comparison of Semantic Label Transfer} over the 10 scenes on KITTI-360.}
\vspace{-0.3cm}
\label{semantic_prediction}
\end{table*}

\section{Experiments} \label{experiment}
\boldparagraph{Dataset} %
We conduct experiments on the recently released 
KITTI-360~\cite{liao2021kitti} dataset. KITTI-360 is collected in suburban areas and provides 3D bounding primitives covering the full scene. 
Following~\cite{liao2021kitti}, we evaluate Panoptic NeRF on manually annotated frames from 5 static suburbs. We split these 5 suburbs into 10 scenes, comprising 128 consecutive frames each with an average travel distance of 0.8m between frames. We leverage all 128 pairs of posed stereo images for training. KITTI-360 provides a set of manually labeled frames sampled in equidistant steps of 5 frames. We use half of the manually labeled frames for evaluation and provide the other half as input to 2D-to-2D label transfer baselines.
We improve the quality of the manually labeled ground truth which is inaccurate at ambiguous regions, see supplementary for details.

\begin{figure*}[tb]
 \centering
 \newcommand{\mywidth}{0.235\textwidth}
 \setlength\tabcolsep{0.2em}
 \begin{tabular}{cccc}
  \includegraphics[width=\mywidth]{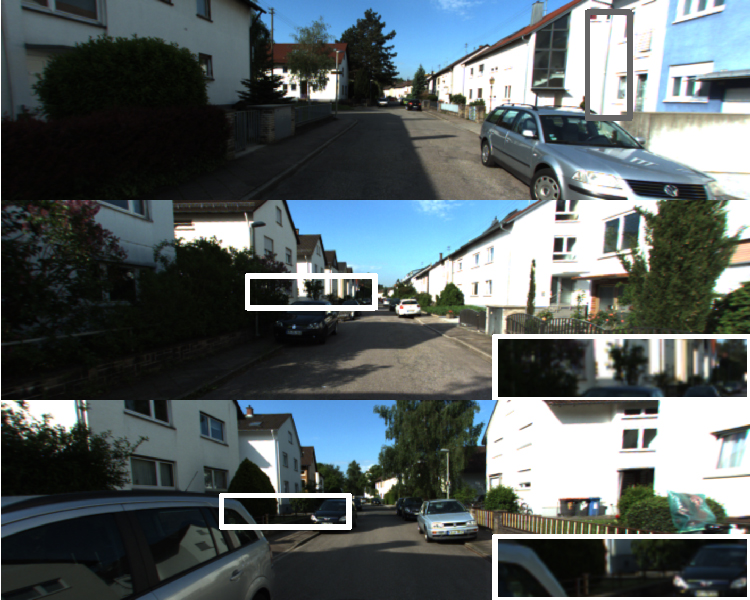}&
  \includegraphics[width=\mywidth]{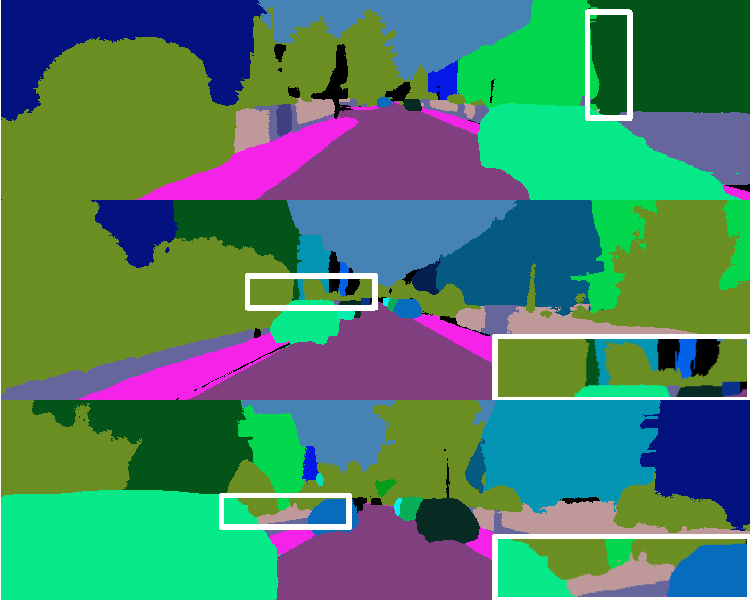} &
  \includegraphics[width=\mywidth]{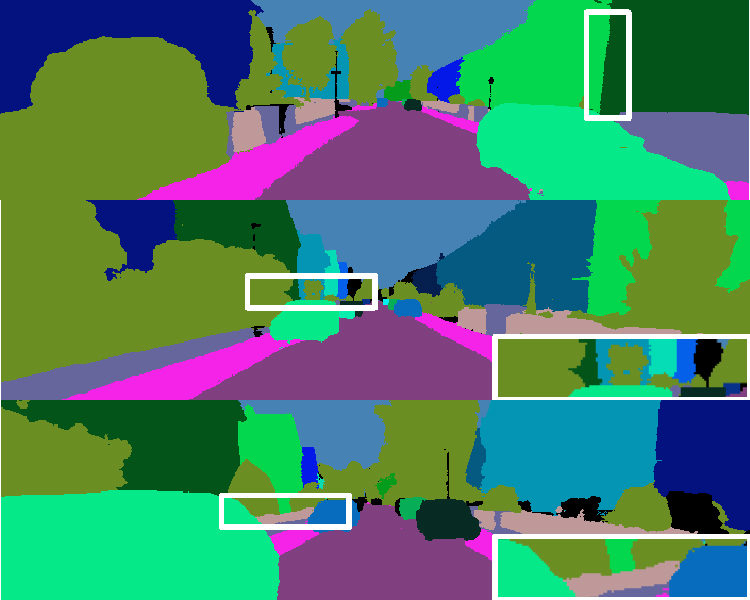}&
  \includegraphics[width=\mywidth]{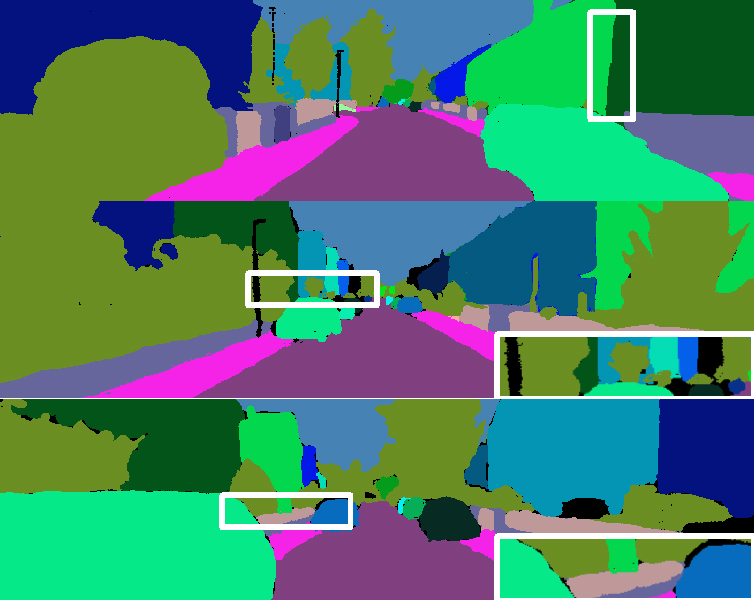}
    \\
    \small{Input} &  
    \small{3D-2D CRF~\cite{liao2021kitti}} &  
     \small{Ours} & 
     \small{GT Label}
 \end{tabular}
 \vspace{-0.3cm}
 \caption{
  \textbf{Qualitative Comparison of Panoptic Label Transfer.} Our method is capable of distinguishing instances based on inferring in 3D space. In contrast, 3D-2D CRF struggles at far and overexposed areas.
 }
 \vspace{-0.3cm}
 \label{fig:pq}
\end{figure*}

\boldparagraph{Baselines} We compare against top-performing baselines in two categories:
(1) \textit{2D-to-2D label transfer baselines}, including Fully Connected CRF (FC CRF)~\cite{krahenbuhl2011efficient} and Semantic NeRF~\cite{zhi2021place}. For both baselines, we provide manually annotated 2D frames as input, sparsely sampled at equidistant steps of 10 frames. Note that labeling these 2D frames takes similar or longer compared to annotating 3D bounding primitives~\cite{xie2016semantic}. As these 2D annotations are extremely sparse, we further provide the same pseudo-2D labels used in our method to Semantic NeRF. (2) \textit{3D-to-2D label transfer baselines}, including PSPNet*,  3D Primitives/Meshs/Points+GC~\cite{liao2021kitti}, and 3D-2D CRF~\cite{liao2021kitti}. All these baselines leverage the same 3D bounding primitives to transfer labels to 2D. Here, PSPNet* is considered 3D-to-2D as it is pre-trained on Cityscapes and fine-tuned on KITTI-360 based on the 3D sparse label projections. The second set of baselines first project 3D primitives/meshes/points to 2D and then apply Graph Cut to densify the label. The 3D-2D CRF densely connects 2D image pixels and 3D LiDAR points, performing inference jointly on these two fields with a set of consistency constraints. 

\begin{table}[tb]
\centering
\resizebox{.48\textwidth}{!} {
\begin{tabular}{|cc|c c c c|cc| c c c c|}
\hline
\multicolumn{2}{|c|}{Method}                      & PQ            & SQ            & RQ & PQ†    & \multicolumn{2}{c|}{Method}                        & PQ            & SQ            & RQ  & PQ†           \\ \hline
\multicolumn{1}{|c|}{\multirow{3}{*}{3D-2D CRF}}  & All  & 62.2          & 79.1          & 76.9    & 64.9     & \multicolumn{1}{c|}{\multirow{3}{*}{Ours}} & All   & \textbf{64.4} & \textbf{79.3} & \textbf{79.6} & \textbf{66.9}    \\
\multicolumn{1}{|c|}{}                      & Things & 60.7          & 79.5          & \textbf{75.2} & 60.7  & 
\multicolumn{1}{c|}{}                      & Things & \textbf{61.9} & \textbf{80.7} & \textbf{75.2} & \textbf{61.9}
\\ 
\multicolumn{1}{|c|}{}                      & Stuff & 63.0          & \textbf{78.9}        & 77.9  & 67.3  &   
\multicolumn{1}{c|}{}                      & Stuff   & \textbf{65.8} & 78.6 & \textbf{82.0} & \textbf{69.8} \\           \hline
\end{tabular}
}
\vspace{-0.2cm}
\caption{\textbf{Quantitative Comparison of Panoptic Label Transfer} over all 10 test scenes on KITTI-360.}
\vspace{-0.2cm}
\label{pq}
\end{table}

\boldparagraph{Pseudo 2D GT}  We use PSPNet* to provide pseudo ground truth in our main experiment to supervise our dual semantic fields. This ensures fair comparison to the 3D-2D CRF, which takes the predictions of PSPNet* as unary terms. 
Note that PSPNet* is fine-tuned on KITTI-360.
To further simplify the entire process, in the ablation study we investigate the performance of our method using pre-trained models on Cityscape without any fine-tuning, including PSPNet~\cite{zhao2017pyramid}, Deeplab~\cite{chen2017deeplab} and Tao \textit{et al.}~\cite{tao2020hierarchical}.

\boldparagraph{Metrics} We evaluate semantic labels by the mean Intersection over Union (mIoU) and the average pixel accuracy (Acc) metrics. To quantitatively evaluate multi-view consistency (MC), we utilize LiDAR points to retrieve corresponding pixel pairs between two consecutive evaluation frames. The MC metric is then calculated as the ratio of pixels pairs with consistent semantic labels over all pairs. For evaluating panoptic segmentation, we report Panoptic Quality (PQ)~\cite{kirillov2019panoptic}, which can be decomposed into Segmentation Quality (SQ) and Recognition Quality (RQ). We additionally adopt PQ†~\cite{porzi2019seamless} as PQ over-penalizes errors of stuff classes.
To verify that Panoptic NeRF is able to improve the underlying geometry, we further evaluate the rendered depth compared to sparse depth maps obtained from LiDAR
using Root Mean Squared Error (RMSE) and the ratio of accurate predictions ($\delta_{1.25}$)~\cite{bhoi2019monocular,fu2018deep}.

\subsection{Label Transfer}
We evaluate our model and compare it to our baselines on KITTI-360.
As most baselines are not designed for panoptic label transfer, we first compare the semantic predictions of all methods and then compare our panoptic predictions to the 3D-2D CRF. 

\boldparagraph{Semantic Label Transfer} As evidenced by \tabref{semantic_prediction}, our method achieves the highest mIoU and Acc over a line of baselines. Specifically, compared to 3D-2D CRF, we obtain an absolute improvement of $1.6\%~ (79.5\% \rightarrow 81.1\%)$ on mIoU. 
Despite PSPNet* being finetuned on KITTI-360 which reduces the performance gap, our method outperforms PSPNet* by a significant margin. Supervised by the extremely sparse manually annotated GT, Semantic NeRF struggles to produce reliable performance. Using pseudo labels of PSPNet*, Semantic NeRF is capable of denoising and thus improving performance $(67.2\% \rightarrow 68.6\%)$. However, both variants of Semantic NeRF are inferior in the urban scenario when the input views are sparse. Moreover, in terms of MC, our method slightly outperforms the 3D-2D CRF and significantly surpasses 2D-to-2D label transfer methods. While our method slightly lags behind 3D Point + GC in terms of MC, it is reasonable as the label consistency is evaluated on the sparsely projected 3D points which GC takes as input to generate a dense label map.

\begin{figure*}[tb]
 \centering
 \newcommand{\mywidth}{0.235\textwidth}
 \setlength\tabcolsep{0.2em}
 \begin{tabular}{cccc}

   \includegraphics[width=\mywidth]{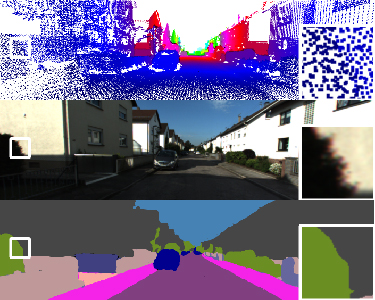}&
  \includegraphics[width=\mywidth]{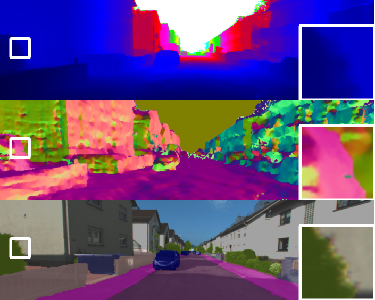}&
  \includegraphics[width=\mywidth]{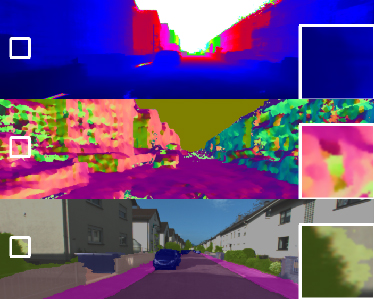}&
  \includegraphics[width=\mywidth]{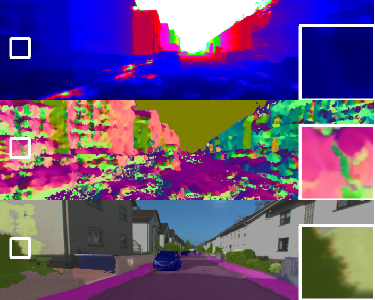}
  \\
     \small{GT}&
     \small{Complete}  &  
     \small{w/o $\cL^{\text{2D}}_{\hat{\bS}}$}& 
     \small{NeRF*} 
 \end{tabular}
 \vspace{-0.3cm}
 \caption{
  \textbf{Ablation Study.} Top: LiDAR depth map (visually enhanced) and rendered depth maps. Middle:  Normal maps obtained from depth maps. Bottom: Semantic GT and predictions. Note that removing $\cL^{\text{2D}}_{\hat{\bS}}$ leads to over-smooth object boundaries and inaccurate semantic segmentation.  
 }
 \label{fig:depth}
 \vspace{-0.45cm}
\end{figure*}

\boldparagraph{Panoptic Label Transfer}
We split the instance labels into things and stuff classes in KITTI-360. 
As the class ``building" is classified as thing in KITTI-360 but stuff in Cityscapes, our 2D baselines~\cite{cheng2020panoptic} are not suitable for testing performance on KITTI-360. Therefore, we ignore pre-trained 2D SOTA baselines. As shown in \tabref{pq}, our proposed method outperforms the 3D-2D CRF in both things and stuff classes. A visual comparison is shown in \figref{fig:pq}. As can be seen, we can deal well with overexposure at buildings, which is a challenge for the 3D-2D CRF, as it projects LiDAR points to reconstruct the intermediate meshes whose quality suffers on building class. 

\boldparagraph{Novel View Label Synthesis}
Panoptic NeRF can render RGB images and panoptic labels at novel viewpoints, whereas 3D-2D CRF is not capable of doing so. Please refer to the supplementary material for details.

\subsection{Ablation Study}
\label{sec:ablation}
We validate our pipeline's design modules with an extensive ablation in \tabref{ablation} by removing one component at a time. As there is a positive correlation between semantic labels and panoptic labels, we focus on semantic segmentation for this experiment on one scene.
\begin{table}[tb!]
\centering
\resizebox{.46\textwidth}{!} 
{

\begin{tabular}{|c|cc|c|cc|} \hline  
& \multicolumn{2}{c|}{Depth(0-100m)} &Evaluated & \multicolumn{2}{c|}{Semantic}\\
&RMSE$\downarrow$ &$\delta_{1.25}\uparrow$ &Label &mIoU &Acc\\ \hline
3D-2D CRF& - & - & - & 79.4 & 93.7  \\
\hline
NeRF* & 10.71 & 78.4&${\hat{\bS}}$ (fixed $s_\beta$) & 67.2 & 88.6 \\
w/o $\cL^{\text{2D}}_{\hat{\bS}}$& 6.28 & 93.1&$\bS$  (learned $s_\phi$) &76.9 & 93.1\\
w/o $\cL_{d}$& 6.15 & 94.0 &$\bS$ (learned $s_\phi$) & 79.1 & 94.1 \\
Uniform S. & 6.28 & 91.1&$\bS$ (learned $s_\phi$) & 73.5 & 92.2 \\
\hline
w/o $\cL^{\text{2D}}_{\bS}$& 6.01 & 95.0&$\bS$ (learned $s_\phi$) & 80.8 & 94.2 \\
w/o $\cL^{\text{2D}}_{\bS}$&6.01 & 95.0& $\hat{\bS}$ (fixed $s_\beta$) & 79.4 & 94.0 \\
w/o $\cL^{\text{3D}}_{\bs}$& 5.74 & 95.0&$\bS$ (learned $s_\phi$) & 70.7 & 92.8 \\
w/o $\bu(\mathbf{r})$& 5.84 & 94.8&$\bS$ (learned $s_\phi$) & 80.8 & 94.4 \\
\hline
Complete & $\mathbf{5.23}$ & $\mathbf{95.1}$&$\bS$ (learned $s_\phi$) & \textbf{81.4} & \textbf{94.5}   \\ \hline 
\end{tabular}
}
\vspace{-0.2cm}
\caption{\textbf{Ablation Study} over 1 test scene on KITTI-360.} 
\vspace{-0.3cm}
\label{ablation}
\end{table}

\boldparagraph{Geometric Reconstruction} We now verify that our method effectively improves the underlying geometry leveraging semantic information. We first remove all the other losses except for $\cL_{p}$, leading to a baseline similar to NeRF but uses our proposed sampling strategy (NeRF*). In this case we render a semantic map based on the fixed semantic field $s_\beta$. As can be seen from  \tabref{ablation} and \figref{fig:depth}, the underlying geometry of NeRF* drops significantly with only $\cL_{p}$. More importantly, the depth prediction also degrades considerably when removing $\cL^{\text{2D}}_{\hat{\bS}}$ (w/o $\cL^{\text{2D}}_{\hat{\bS}}$), indicating the importance of the fixed semantic field in improving the underlying geometry. \figref{fig:depth} shows that the full model has sharper edges while removing the fixed semantic fields leads to over-smooth object boundaries. We further show that eliminating $\cL_d$ (w/o $\cL_d$) or replacing the sampling strategy with standard uniform sampling (Uniform S.) both impair the geometric reconstruction, and consequently the 
semantic estimation as well. 

\boldparagraph{Semantic Segmentation} When removing $\cL^{\text{2D}}_{\bS}$ (w/o $\cL^{\text{2D}}_{\bS}$), the performance also drops as the learned semantic field $s_\phi$ is only supervised by weak 3D supervision. Interestingly, this baseline still outperforms the semantic map rendered by the fixed semantic field despite that they share the same geometry. This observation suggests that the weak 3D supervision provided by $\cL^{\text{3D}}_{\bs}$ also allows to address the label ambiguity of the overlapping region to a certain extent. Therefore, it is not surprising that removing $\cL^{\text{3D}}_{\bs}$ (w/o $\cL^{\text{3D}}_{\bs}$) worsens the semantic prediction compared to the full model. 
We observe that the performance is also deteriorated without ray masking (w/o $\bu(\br)$).

\boldparagraph{2D Pseudo GT}Finally, 
we evaluate how the quality of the pseudo 2D ground truth affects our method in \tabref{tab4}. 
As some classes are not considered during training in Cityscapes, we additionally report mIoU$_\text{sub}$ over the remaining classes. It is worth noting that using models pre-trained on Cityscapes without any fine-tuning leads to promising results, where  Ours w/ Deeplab~\cite{chen2017deeplab} and Ours w/ Tao \etal~\cite{tao2020hierarchical} are very close to Ours + PSPNet* in terms of mIoU$_\text{sub}$. More importantly, our method consistently outperforms the corresponding pseudo GT leveraging the 3D bounding primitives.

\begin{table}[tb]
\centering
\resizebox{.48\textwidth}{!} {

\begin{tabular}{|c|ccc|c|ccc|} \hline  
Method &mIoU &mIoU$_\text{sub}$ &Acc &Method &mIoU &mIoU$_\text{sub}$ &Acc \\  \hline
Deeplab ~\cite{chen2017deeplab} &  - &74.7 & 88.1 & Ours w/ ~\cite{chen2017deeplab} & 79.3& 86.3 & 94.0\\
Tao \textit{et al.} ~\cite{tao2020hierarchical}& - &79.9 & 91.2 & Ours w/ ~\cite{tao2020hierarchical} & 79.8 & 86.1 & 94.5 \\
PSPNet~\cite{zhao2017pyramid} & - &70.9 & 87.0 & Ours w/ \cite{zhao2017pyramid} & 75.9 & 81.7 &91.6\\
PSPNet*~\cite{zhao2017pyramid} & 62.0 & 77.5 & 90.1 & Ours & 81.4  & 87.3 & 94.5\\

\hline
\end{tabular}
}
\vspace{-0.2cm}
\caption{\textbf{Quantitative Comparison} using different 2D Pseudo GTs.}
\vspace{-0.3cm}
\label{tab4}
\end{table}

\subsection{Limitations}
Our method performs per-scene optimization and  training takes 4 hours on one scene. Training time needs to be reduced to scale our method to large-scale scenes, \eg, by adopting improvements in speeding up NeRF training~\cite{muller2022instant,yu2021plenoxels}. In addition, we consider label transfer on static scenes in this work. We plan to extend our method to dynamic scenes leveraging recent advances in dynamic radiance field estimation~\cite{Ost2021CVPR}.

\section{Conclusion} 
We present Panoptic NeRF that infers in 3D space and renders per-pixel semantic and instance labels for 3D-to-2D label transfer. By combining coarse 3D bounding primitives and noisy 2D predictions using our dual semantic fields, Panoptic NeRF is capable of improving the underlying geometry given sparse input views and resolving label noise. Moreover, it enables label synthesis at novel view points. We believe that our method is a step towards more efficient data annotation, while simultaneously providing a 3D consistent continuous panoptic representation of the scene.

\clearpage
{\small
\bibliographystyle{ieee_fullname}
\bibliography{bibliography,bibliography_long,bibliography_custom}
}

\clearpage
\onecolumn
\appendix

\section{Implementation Details}\label{implementation}

\subsection{Network Architecture}
\begin{figure}[htb]
\centerline{\includegraphics[width=.7\textwidth]{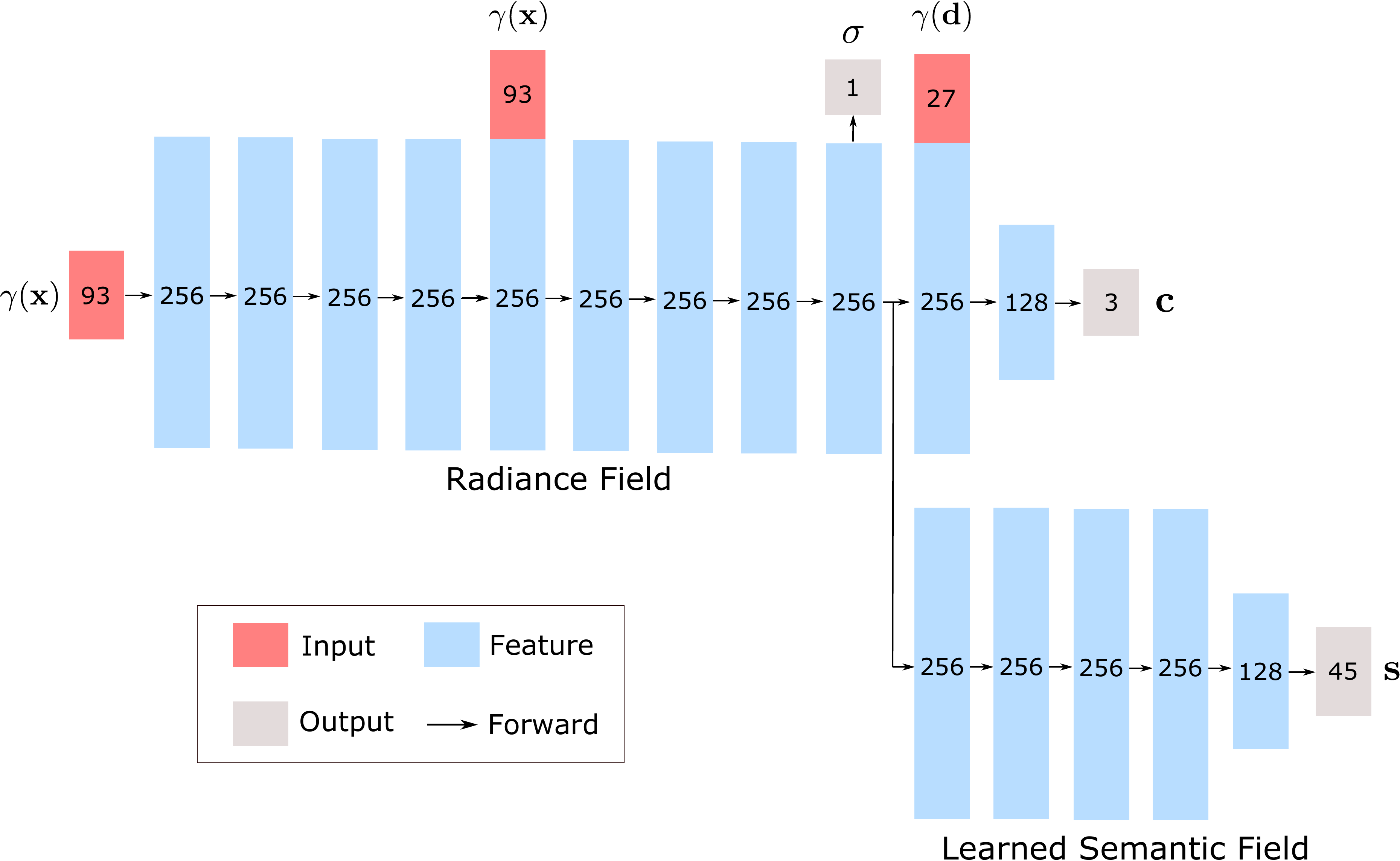}}
\caption{\textbf{Trainable Part of Panoptic NeRF.} For the radiance field, we follow the original implementation of NeRF except for setting $L=15$ for $\gamma(\bx)$. The learned semantic field predicts semantic logits independent of the viewing direction.
The logits are then transformed into categorical distributions through a softmax layer.
}
\vspace{-1.5em}
\label{fig:network_supp}
\end{figure}

\figref{fig:network_supp} shows the trainable part of our Panoptic NeRF model. We adopt the same network architecture in all experiments. 
The network takes as input the 3D location $\mathbf{x}$ (each element normalized to $[-1,1]$) and the viewing direction $\mathbf{d}$. Following NeRF~\cite{mildenhall2020nerf}, both $\bx$ and $\bd$ are mapped to a higher dimensional space using a positional encoding (PE):
\begin{equation}
\gamma(p)=(\sin (2^{0} \pi p), \cos (2^{0} \pi p), \cdots, \sin (2^{L-1} \pi p), \cos (2^{L-1} \pi p))
\end{equation}

To learn high frequency components in unbounded outdoor environments, we set $L=15$ for $\gamma(\bx)$ and $L=4$ for $\gamma(\bd)$. 
Our learned semantic field  is conditioned only on the 3D location $\bx$ rather than the viewing direction $\bd$ in order to predict view-independent semantic distributions.

\subsection{Sampling Strategy}
We sample points within the bounding primitives to skip empty space. As our bounding primitives are convex\footnote{The cuboids and ellipsoid are both convex. The extruded 3D plane is convex in a local region.}, each ray intersects a bounding primitive exactly twice which determines the sampling interval.
For each camera ray, we sort all bounding primitives that the ray hits from near to far and save the intersections offline.
To save storage and to speed up training, we keep the first 10 sorted bounding primitives as the rest are highly likely to be occluded.  If a camera ray intersects less than 10 bounding primitives, we additionally sample a set of points to model the sky in $[t_{max}, t_{max} + t_{int}]$, where $t_{max}$ denotes the distance from the origin to the furthest bounding primitive in the scene and $t_{int}$ is a constant distance interval.

\subsection{Evaluation Metric}
We evaluate mIoU and pixel accuracy following standard practice~\cite{cordts2016cityscapes,liao2021kitti}. Here, we provide more details of the multi-view consistency and panoptic quality metrics.

\boldparagraph{Multi-view Consistency} To evaluate multi-view consistency, we use depth maps obtained from LiDAR points to retrieve matching pixels across two consecutive frames. A similar multi-view consistency metric is considered in~\cite{Tong2021IJCAI} where optical flow is used to find  corresponding pixel pairs. We instead use LiDAR depth maps as they are more accurate compared to optical flow estimations. The details of generating the LiDAR depth maps will be introduced in \secref{sec:lidar_depth}. Given LiDAR depth maps at two consecutive test frames, we first unproject them into 3D space and find matching points. Two LiDAR points are considered matched if their distance in 3D is smaller than 0.1 meters. For each pair of matched points, we retrieve the corresponding 2D semantic labels and evaluate their consistency. The MC metric is evaluated as the number of consistent pairs over all matched pairs. Despite being not 100$\%$ accurate as the 3D points may not match exactly in 3D space, we find this metric meaningful in reflecting multi-view consistency.

\boldparagraph{Panoptic Quality}
Following~\cite{kirillov2019panoptic}, we use the PQ metric to evaluate the performance of panoptic segmentation. 
As the ground truth panoptic labels are not precise in distant areas and have a lot of small noises of things, we set ground truth labels of areas less than 100 pixels to ``void''. Correspondingly, segment matching will not be performed in void regions. In addition, Panoptic maps of the 3D-2D CRF and our method are obtained by 3D primitives, thus containing very far objects. In fact, these far objects may only occupy very small areas, usually less than 100 pixels, on 2d images. To avoid being biased by those extremely far objects in the segment matching, we omit them by setting the predicted labels of the areas less than 100 pixels to the ``sky'' class.  To ensure a fair comparison across all methods, we adopt the same evaluation protocol for all baselines and our method.

\subsection{Training and Inference}
As mentioned in Section 4.2 of the main paper, our total loss function   comprises five terms, including three semantic losses $\cL^{2D}_{\hat{\bS}}$, $\cL^{2D}_{\bS}$, $\cL^{3D}_{\bs}$, the photometric loss $\cL_{\bc}$ and the depth loss $\cL_{d}$. 
During per-scene optimization, the photometric loss  $\cL_{\bc}$ is defined on the posed stereo images. The 2D semantic losses $\cL^{2D}_{\hat{\bS}}$, $\cL^{2D}_{\bS}$ are applied to the left images only. While our method allows for using noisy 2D semantic predictions on the right images, this ensures fair comparison to the 3D-2D CRF which is not capable of using predictions on other viewpoints for inference. We apply the 3D semantic loss $\cL^{3D}_{\bs}$ directly on 3D points sampled along the camera rays of the left images. The depth loss $\cL_d$ is also defined on the left images as the information gain is marginal on the right views. 

For inference, we compare our method to the baselines on the left views of which the manually labeled 2D Ground Truth is defined. Note that our method is not constrained to the left views during inference. We show label transfer results on the right camera views in \secref{sec:stereo_view} and novel view label synthesis in \secref{sec:novel_view}.

\section{Data Preparation} \label{datapre}

\subsection{Stereo Depth for Weak Depth Supervision}
\begin{figure*}[tb]
 \centering
 \newcommand{\mywidth}{0.33 \textwidth}
 \setlength\tabcolsep{0.2em}
 \begin{tabular}{ccc}
      \includegraphics[width=\mywidth]{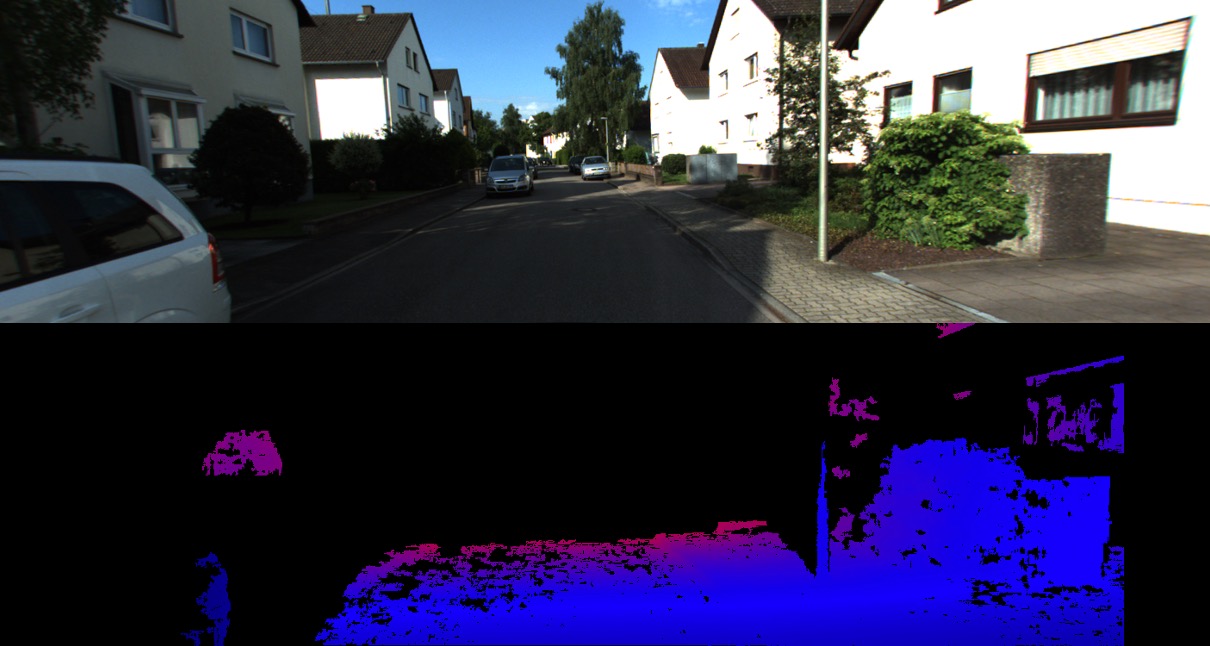}&
  \includegraphics[width=\mywidth]{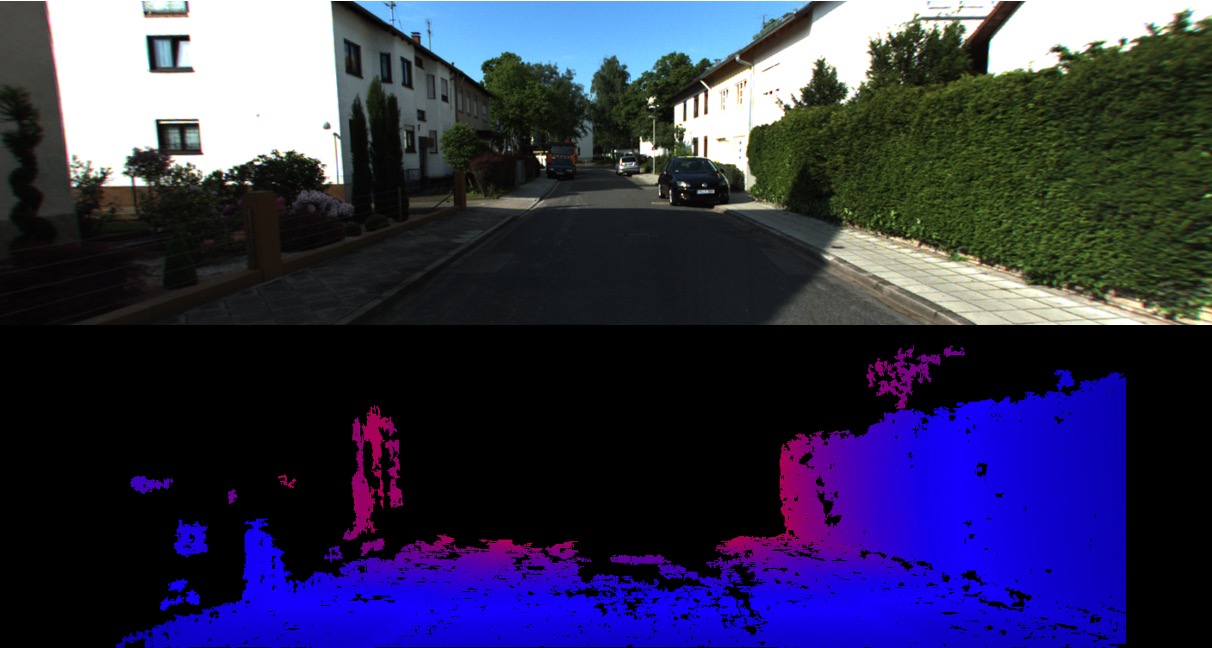}&       \includegraphics[width=\mywidth]{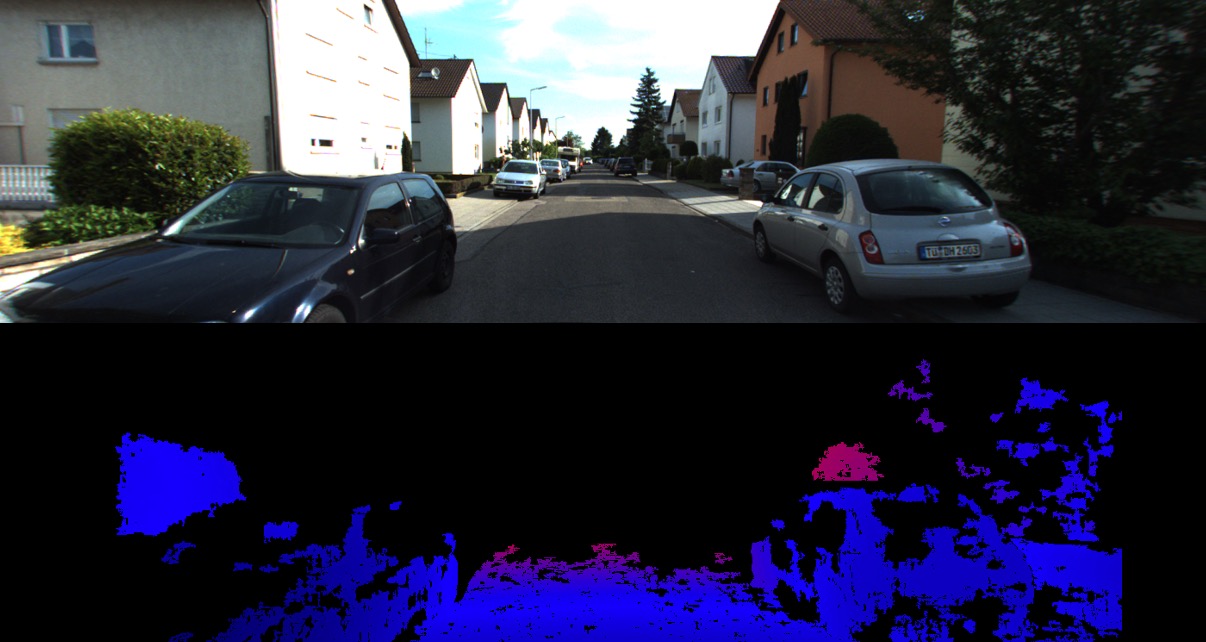}\\
   \includegraphics[width=\mywidth]{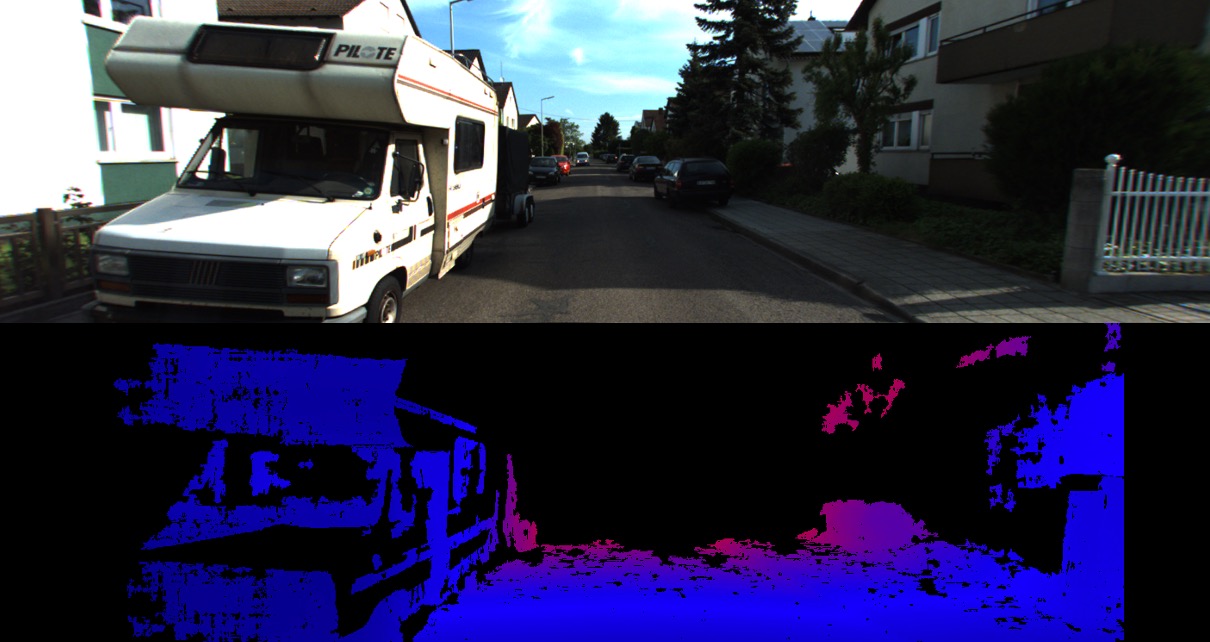}&
  \includegraphics[width=\mywidth]{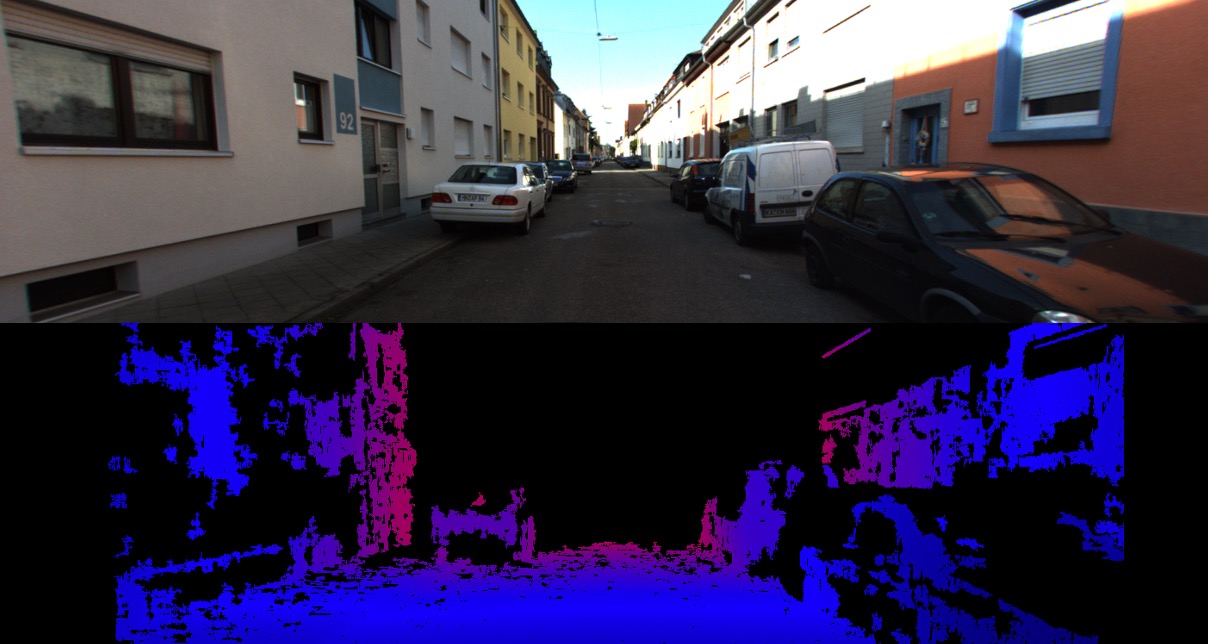}&       \includegraphics[width=\mywidth]{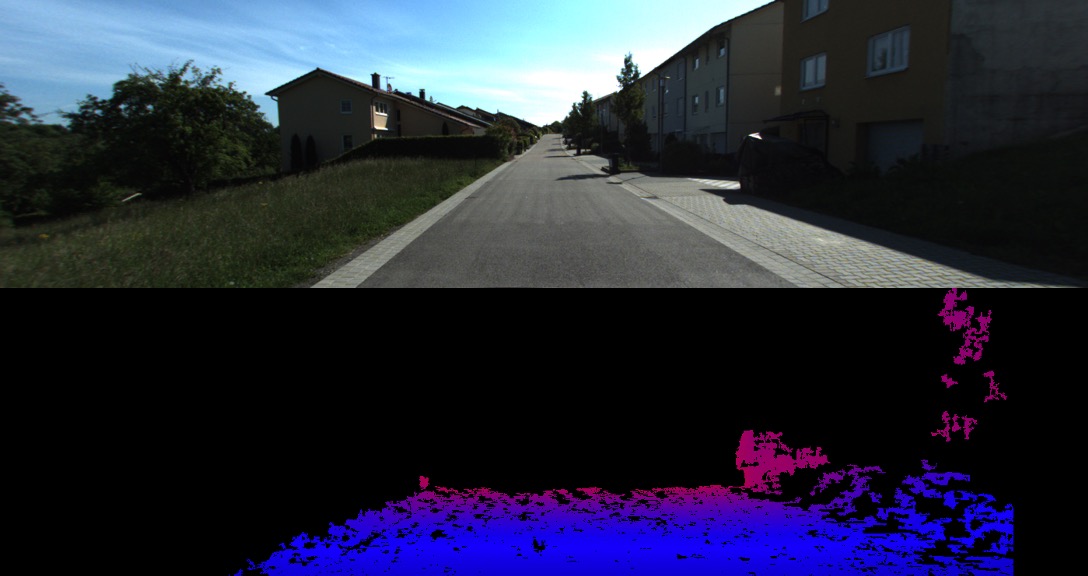}\\

 \end{tabular}
 \caption{
  \textbf{Depth Maps for Weak Depth Supervision.} Each group shows the RGB image (top) and the corresponding depth maps (bottom) used for supervision.}
 \label{fig:sgmdepth}

\end{figure*}

To provide weak depth supervision to Panoptic NeRF, we use Semi-Global Matching (SGM)~\cite{hirschmuller2007stereo} to estimate depth given a stereo image pair. We perform a left-right consistency check and a multi-frame consistency check in a window of $5$ consecutive frames to filter inconsistent predictions. We further omit depth predictions further than 15 meters for each frame as disparity is better estimated in nearby regions, see \figref{fig:sgmdepth}.

\begin{figure*}[tb]
 \centering
 \newcommand{\mywidth}{0.49\textwidth}
 \setlength\tabcolsep{0.8em}
 \begin{tabular}{cc}
  \includegraphics[width=\mywidth]{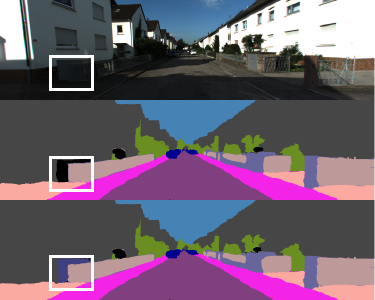}&
  \includegraphics[width=\mywidth]{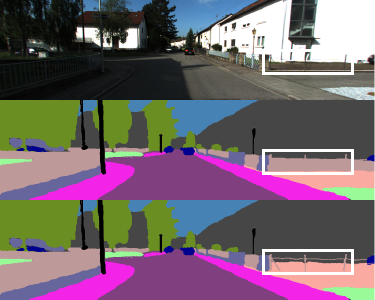}    

 \end{tabular}
 \caption{
  \textbf{Examples of Modified Ground Truth.} We correct some GT pixels that were incorrectly labeled in the KITTI-360 dataset. Top: Input RGB images. Middle: Original ground truth. Bottom: Modified ground truth. In the first column, we add the ``box" class. In the second column, we correct the ``parking" area.
 }
 
 \label{fig:modgt}
\end{figure*}

\subsection{LiDAR Depth for Evaluation}
\label{sec:lidar_depth}
We evaluate the rendered depth maps against the LiDAR measurements. We refrain from using LiDAR as input as 1) this allows us to evaluate our depth prediction against LiDAR and 2) it makes our method more flexible to work with settings without any LiDAR observations. As LiDAR observations at each frame are sparse, we accumulate multiple frames of LiDAR observations and project the visible points to each frame similar to~\cite{uhrig2017sparsity}. 

\subsection{Manually Annotated 2D GT}
The manually annotated 2D ground truth of KITTI-360~\cite{liao2021kitti} is inferior at some regions. For a fair comparison, we improve the label quality by manually relabeling ambiguous classes, see \figref{fig:modgt} for illustrations.

\section{Additional Experimental Results} \label{additionalexper}

\subsection{Weak Depth Supervision}
\begin{figure*}[!t]
 \centering
 \newcommand{\mywidth}{0.95\textwidth}
 \setlength\tabcolsep{0.05em}
 \newcolumntype{P}[1]{>{\centering\arraybackslash}m{#1}}
 \def\arraystretch{0.50}
  \begin{tabular}{P{0.5em}P{0.5em}P{\mywidth}}
    \rot{\scriptsize{Complete Model}}&& \includegraphics[width=\mywidth]{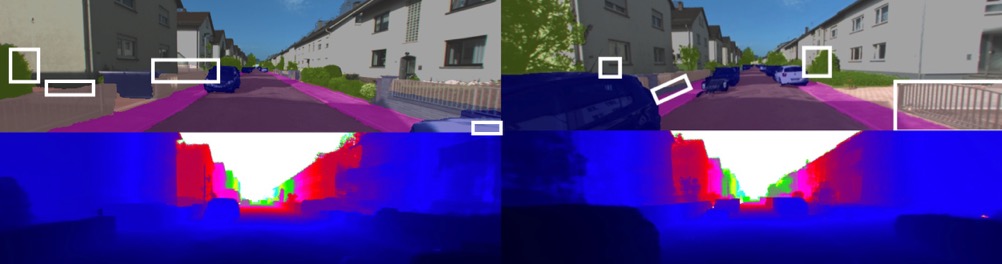}\\
   \rot{\scriptsize{w/ $\cL_d$, w/o $\cL^{2D}_{\hat{\bS}}$}}&& \includegraphics[width=\mywidth]{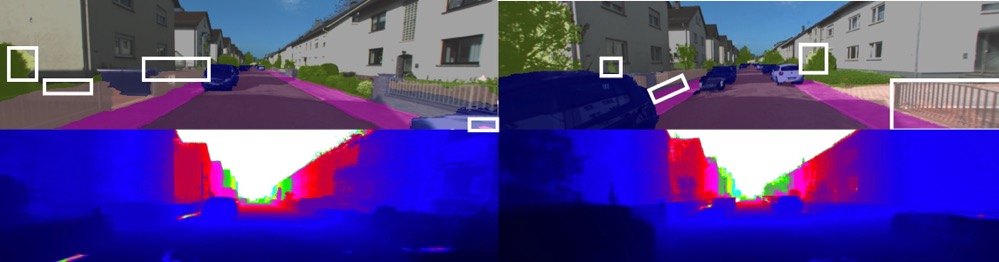}\\
 \end{tabular}
 \vspace{-0.2cm}
 \caption{
  \textbf{Qualitative Comparison of Ablation Study.} We visualize the semantic map and depth map of the complete model (top) and the model without fixed semantic field (bottom). 
 }
 \label{fig:fixablation}

\end{figure*}

We show that using the depth loss $\cL_{d}$ alone is not able to recover accurate object boundaries in \figref{fig:fixablation}.
In contrast, adding the semantic loss $\cL^{2D}_{\hat{\bS}}$ to the fixed semantic field further improves the object boundary. These improvements can be explained as follows: Firstly, the weak stereo depth supervision is not fully accurate, especially at far regions. Furthermore, even with perfect depth supervision, the model receives very small penalty if the predicted depth is close to the GT depth.  In contrast, the cross entropy loss $\cL^{2D}_{\hat{\bS}}$ defined on the fixed semantic field provides a strong penalty as small errors in depth lead to wrong semantics.

\subsection{Qualitative Comparison of Label Transfer}
\begin{figure*}[!t]
 \centering
 \newcommand{\mywidth}{0.85\textwidth}
 \setlength\tabcolsep{0.05em}
 \newcolumntype{P}[1]{>{\centering\arraybackslash}m{#1}}
 \def\arraystretch{0.50}
  \begin{tabular}{P{0.5em}P{0.5em}P{\mywidth}}
    \rot{\tiny{RGB}}&& \includegraphics[width=\mywidth]{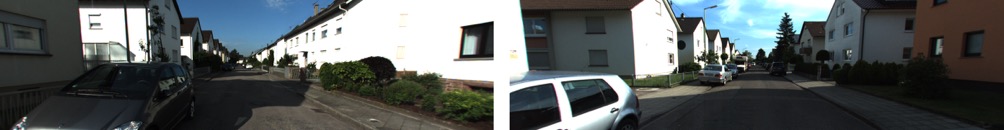}\\
   \rot{\tiny{FC CRF + Manual GT}}&& \includegraphics[width=\mywidth]{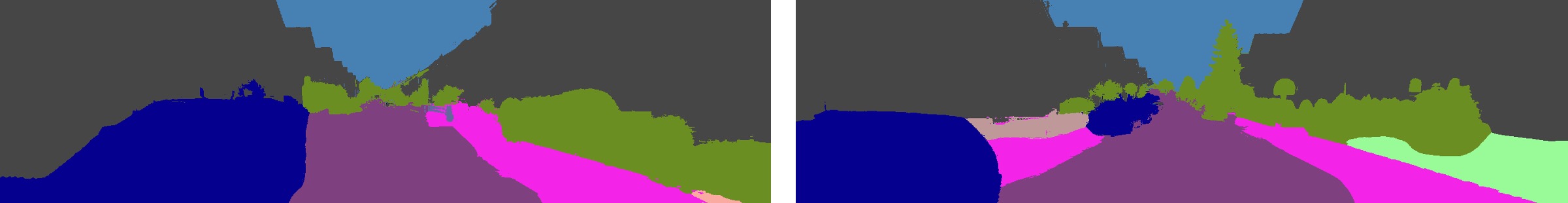}\\
   \rot{\tiny{S-NeRF + Manual GT}}&& \includegraphics[width=\mywidth]{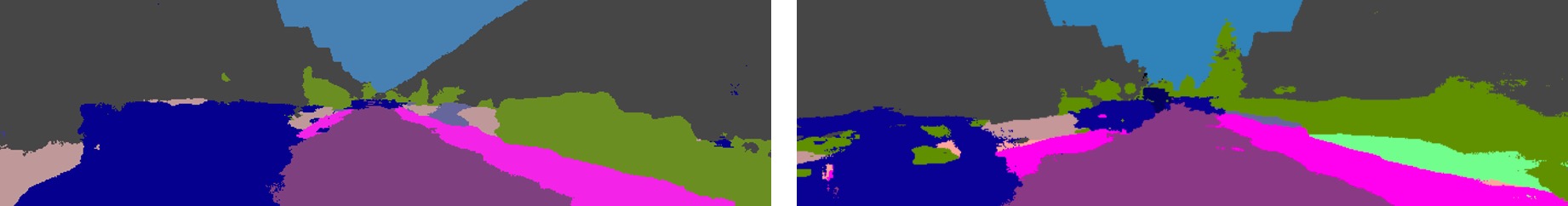}\\
   \rot{\tiny{S-NeRF + PSPNet*}}&& \includegraphics[width=\mywidth]{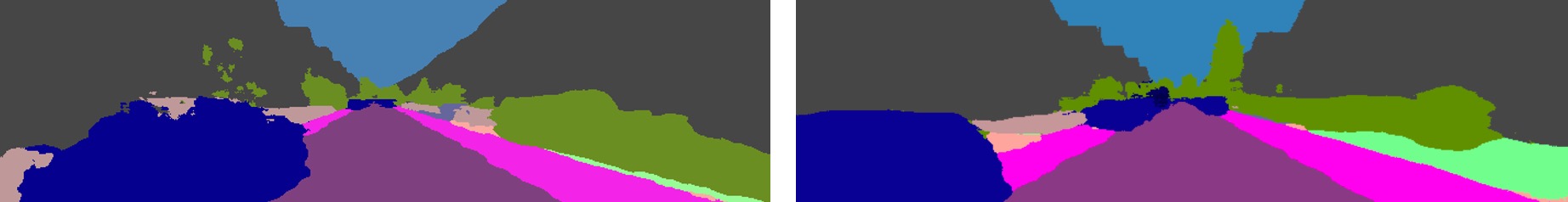}\\
   \rot{\tiny{PSPNet*}}&& \includegraphics[width=\mywidth]{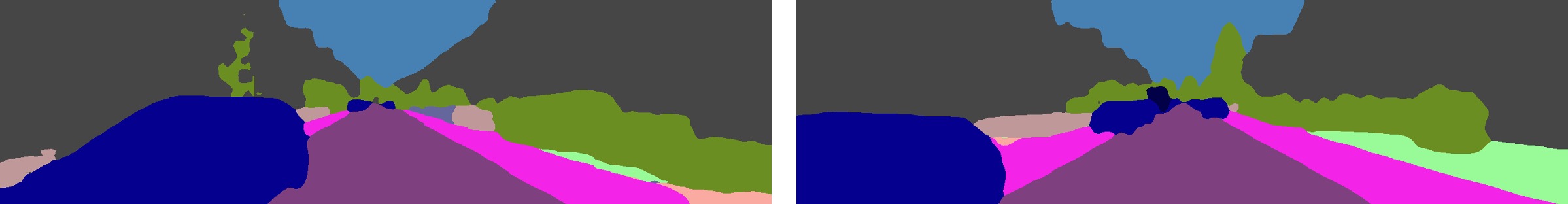}\\
   \rot{\tiny{3D Primitives + GC}}&& \includegraphics[width=\mywidth]{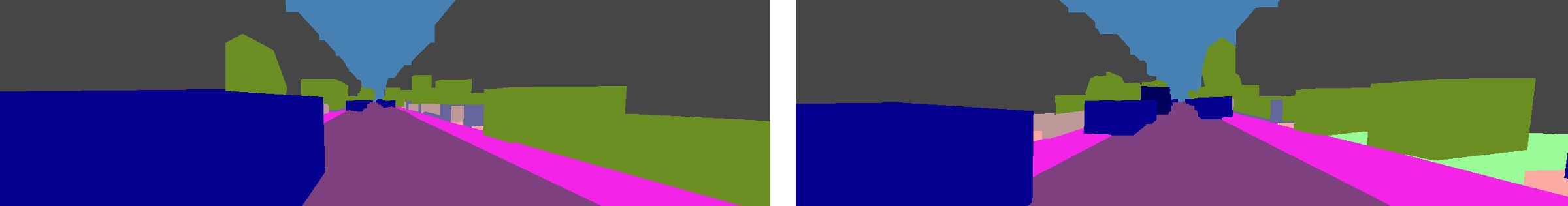}\\
   \rot{\tiny{3D Mesh + GC}}&& \includegraphics[width=\mywidth]{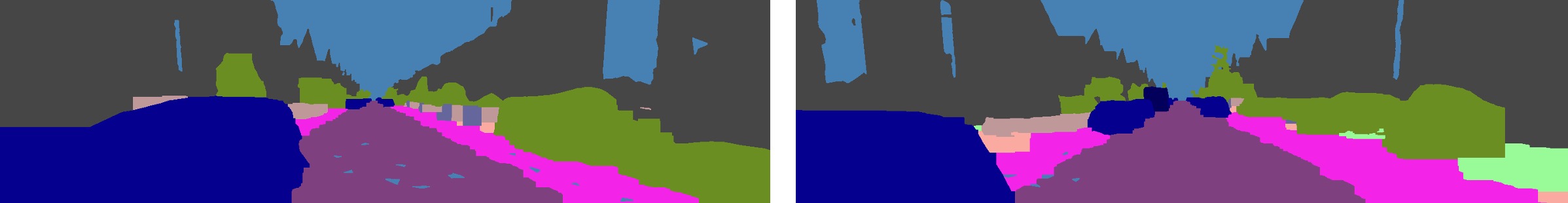}\\
   \rot{\tiny{3D Point + GC}}&& \includegraphics[width=\mywidth]{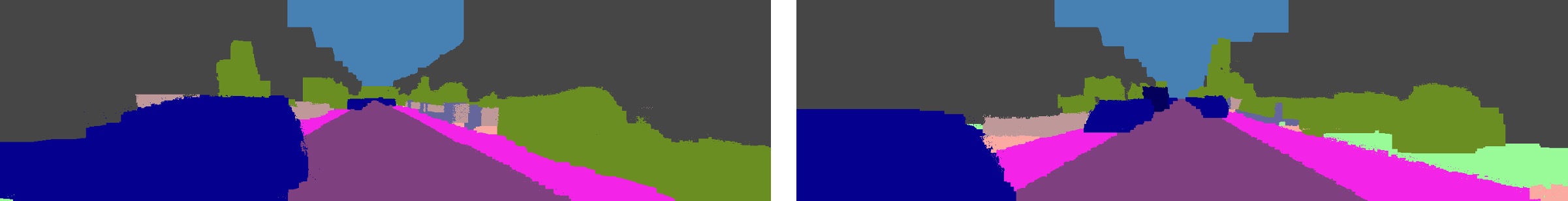}\\
   \rot{\tiny{3D-2D CRF}}&& \includegraphics[width=\mywidth]{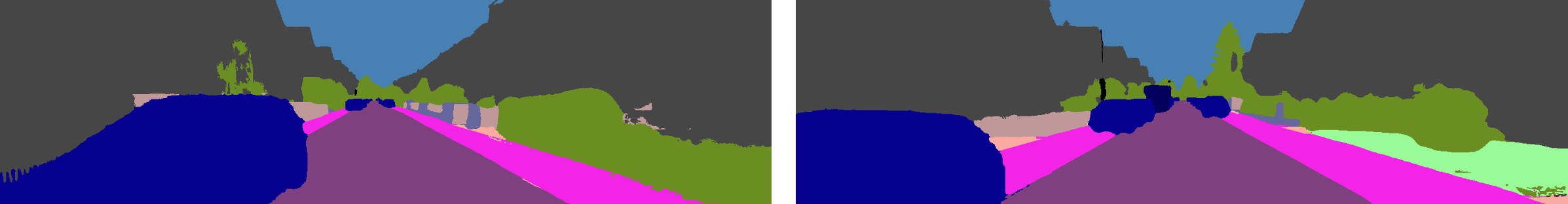}\\
   \rot{\tiny{Ours}}&& \includegraphics[width=\mywidth]{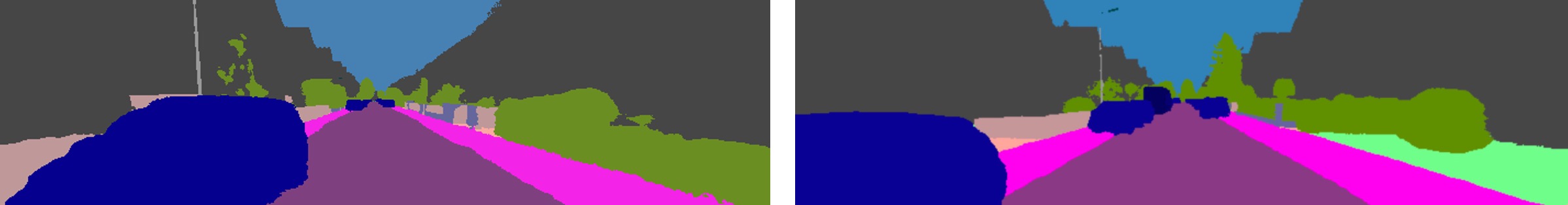}\\
   \rot{\tiny{GT}}&& \includegraphics[width=\mywidth]{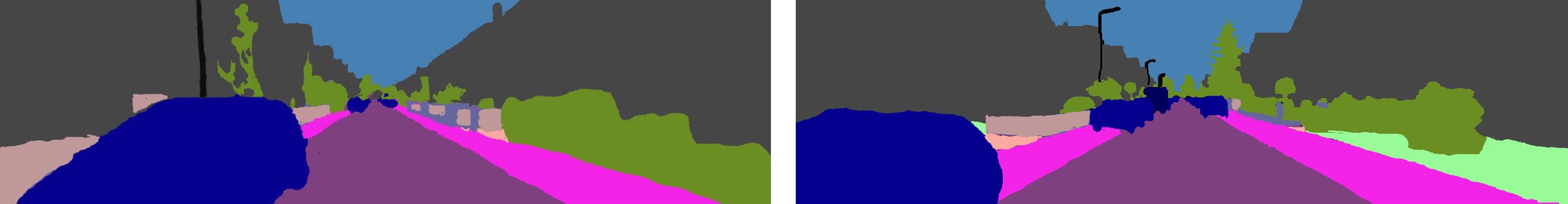}\\
 \end{tabular}
 \caption{
  \textbf{Qualitative Comparison of Semantic Label Transfer} on frames with manually labeled ground truth.
 }
 \label{fig:semanticall}
\end{figure*}

\begin{figure*}[tb]
 \centering
 \newcommand{\mywidth}{0.33 \textwidth}
 \setlength\tabcolsep{0.2em}
 \begin{tabular}{ccc}
      \includegraphics[width=\mywidth]{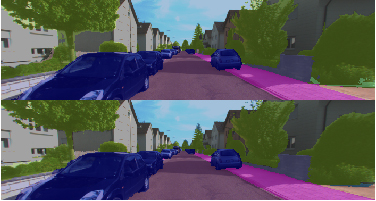}&
  \includegraphics[width=\mywidth]{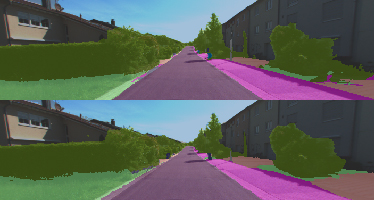}&       \includegraphics[width=\mywidth]{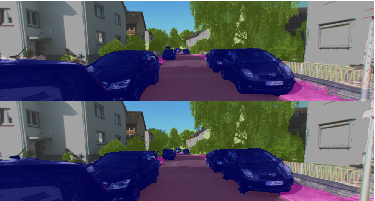}\\
   \includegraphics[width=\mywidth]{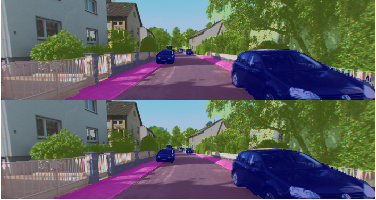}&
  \includegraphics[width=\mywidth]{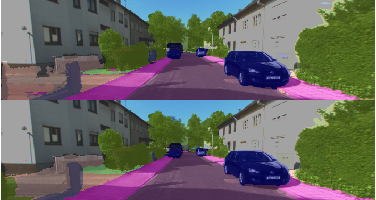}&       \includegraphics[width=\mywidth]{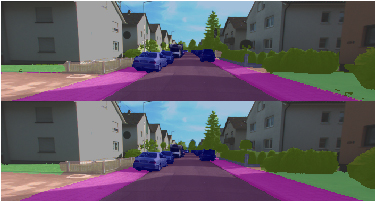}\\

 \end{tabular}
 \vspace{-0.2cm}
 \caption{
  \textbf{Qualitative Comparison of Semantic Label Transfer} on frames without manually labeled ground truth.  Each group shows the prediction of 3D-2D CRF~\cite{liao2021kitti} (top) and ours (bottom).
 }
 \label{fig:moresemanitc}
\end{figure*}

\begin{figure*}[tb]
 \centering
 \newcommand{\mywidth}{0.33 \textwidth}
 \setlength\tabcolsep{0.2em}
 \begin{tabular}{ccc}
      \includegraphics[width=\mywidth]{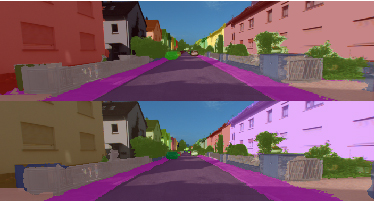}&
  \includegraphics[width=\mywidth]{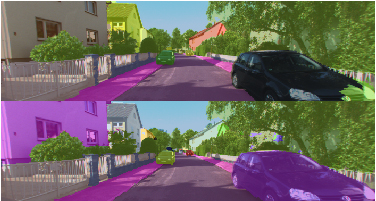}&       \includegraphics[width=\mywidth]{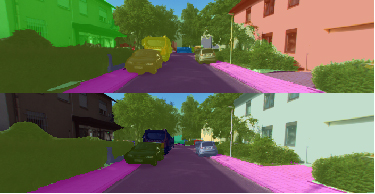}\\
   \includegraphics[width=\mywidth]{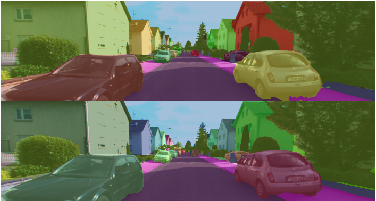}&
  \includegraphics[width=\mywidth]{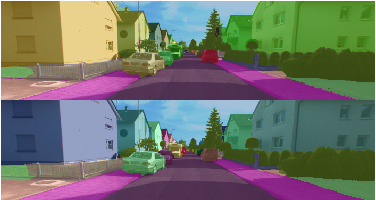}&       \includegraphics[width=\mywidth]{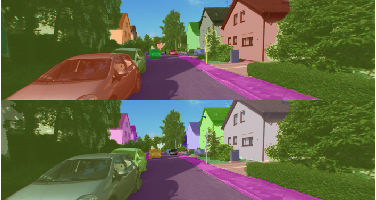}\\
 \end{tabular}
 \vspace{-0.2cm}
 \caption{
  \textbf{Qualitative Comparison of Panoptic Label Transfer} on frames without manually labeled ground truth.  Each group shows the prediction of 3D-2D CRF~\cite{liao2021kitti} (top) and ours (bottom). The colors of the instances do not match due to missing 2D ground truth.
  }
 \label{fig:morepanoptic}
\end{figure*}

\figref{fig:semanticall} shows additional qualitative comparisons corresponding to the Table 1 of the main paper. Consistent with the quantitative results, our method outperforms all baselines qualitatively.
We further show qualitative comparisons to 3D-2D CRF on a set of unlabeled 2D frames, including semantic label transfer in  \figref{fig:moresemanitc} and panoptic label transfer in \figref{fig:morepanoptic}.

\subsection{Stereo Label Transfer}
\label{sec:stereo_view}
\begin{figure*}[tb]
 \centering
 \newcommand{\mywidth}{0.33 \textwidth}
 \setlength\tabcolsep{0.2em}
 \begin{tabular}{ccc}

   \includegraphics[width=\mywidth]{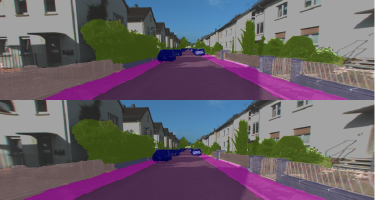}&
  \includegraphics[width=\mywidth]{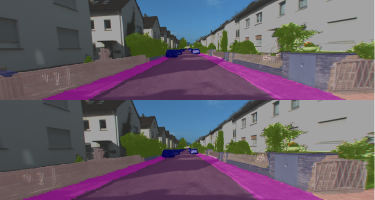}&       \includegraphics[width=\mywidth]{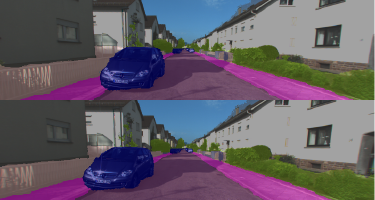}

 \end{tabular}
 \caption{
  \textbf{Qualitative Results for Stereo Label Transfer.} Top: Blended semantic results of left views. Bottom: Blended semantic results of right views.
 }
 \label{fig:trainstereo}
\end{figure*}

In \figref{fig:trainstereo}, we illustrate our stereo label transfer results. Despite that we only utilize pseudo ground truth on the left views for supervision, our model achieves consistent results on both left and right views.

\subsection{Novel View Label Synthesis}
\label{sec:novel_view}
\begin{figure*}[tb]
 \centering
 \newcommand{\mywidth}{0.33 \textwidth}
 \setlength\tabcolsep{0.2em}
 \begin{tabular}{ccc}

   \includegraphics[width=\mywidth]{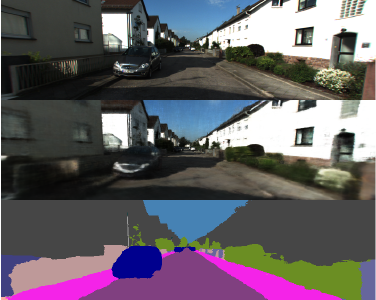}&
  \includegraphics[width=\mywidth]{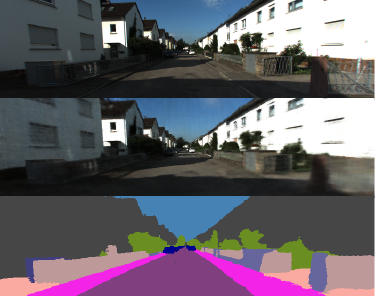}&       \includegraphics[width=\mywidth]{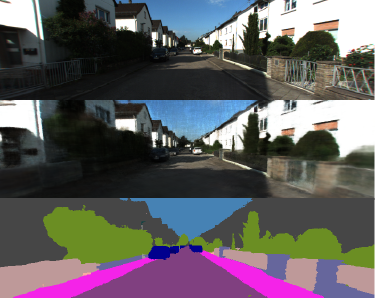}

 \end{tabular}
 \caption{
  \textbf{Qualitative Results for Novel View Label Synthesis.} Top: GT RGB images. Middle: Rendered RGB images. Bottom: Rendered semantic maps. 
 }
 \label{fig:stereo}
\end{figure*}

Here, we evaluate our performance on novel view label synthesis by applying the photometric loss $\cL_{\bc}$ to the left images only. This allows us to evaluate novel view appearance and label synthesis on the right view images.  As shown in \figref{fig:stereo}, our method achieves promising results on novel view appearance and label synthesis. More results of appearance and label synthesis on unseen viewpoints can be found in the supplementary video.

\subsection{Analysis of 3D-2D CRF}
\begin{figure*}[tb]
 \centering
 \newcommand{\mywidth}{0.33 \textwidth}
 \setlength\tabcolsep{0.2em}
 \begin{tabular}{ccc}

   \includegraphics[width=\mywidth]{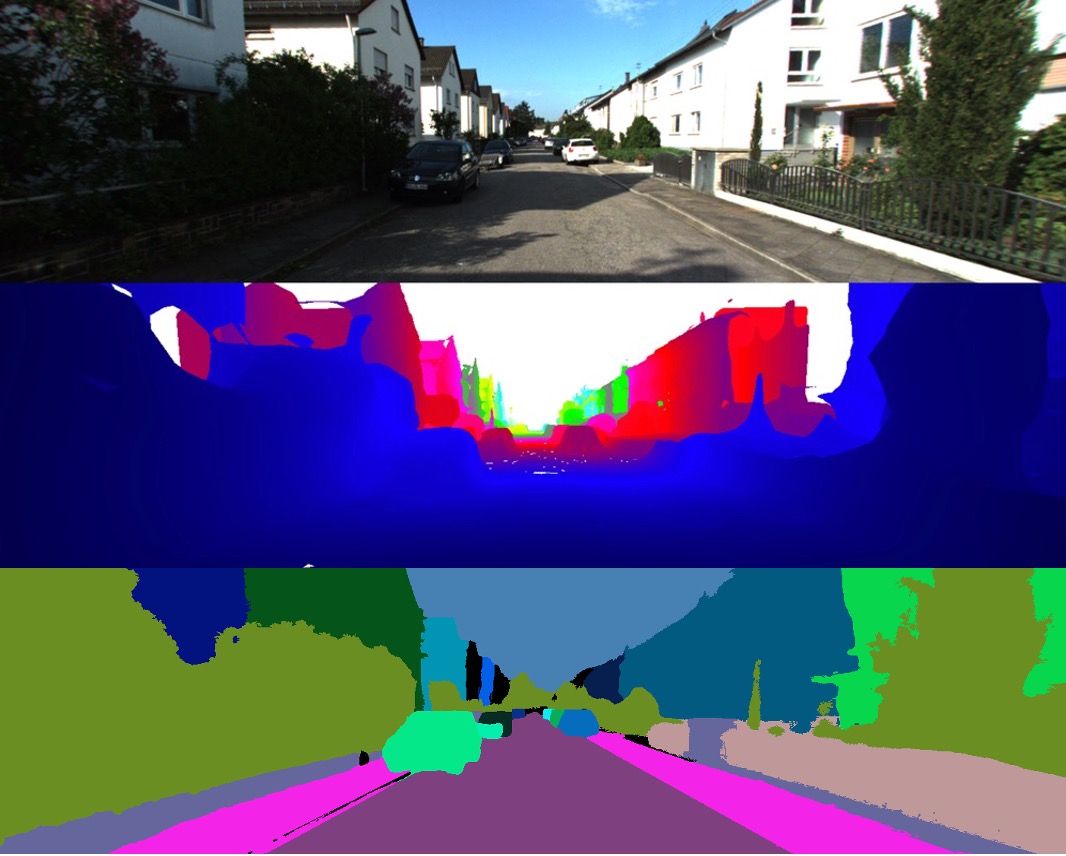}&
  \includegraphics[width=\mywidth]{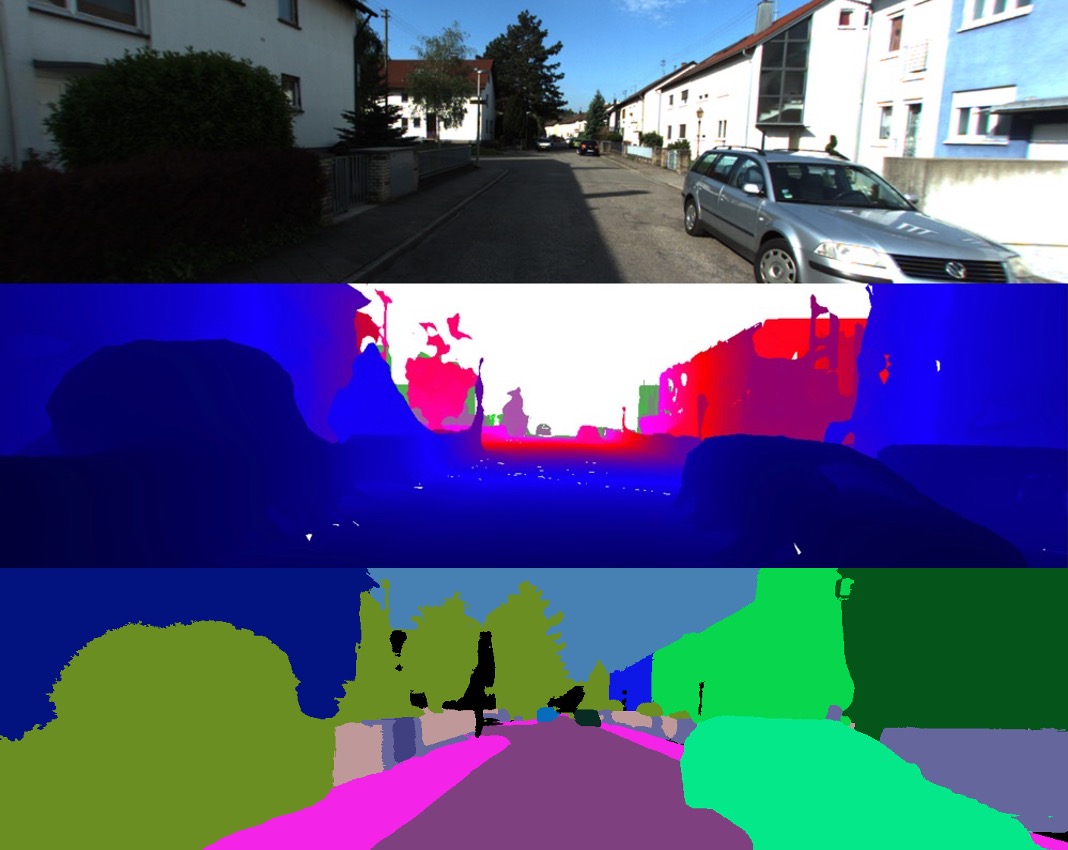}&       \includegraphics[width=\mywidth]{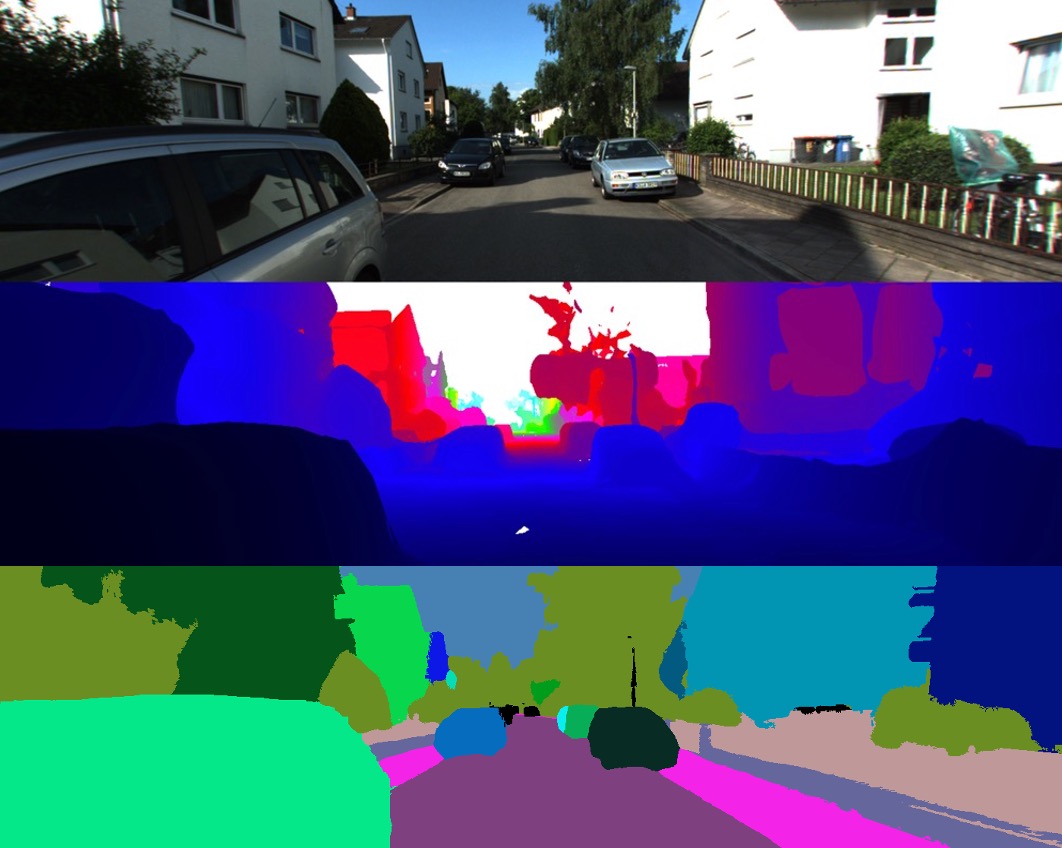}

 \end{tabular}
 \caption{
  \textbf{Qualitative Results of 3D-2D CRF.} Top: Input RGB images. Middle: 3D-2D CRF mesh depth. Bottom: Panoptic label transfer results of the 3D-2D CRF method.}
 \label{fig:meshdepth}
\end{figure*}

The 3D-2D CRF performs inference based on a multi-field CRF which reasons jointly about the labels of the 3D points and all pixels in the image. To obtain dense 3D points, it accumulates LiDAR observations over multiple frames and project visible 3D points to the image based on a reconstructed mesh. \figref{fig:meshdepth} shows depth maps of the reconstructed mesh corresponding to Fig. 5 of the main paper. As can be seen, the side of the building can hardly be scanned by the LiDAR, leading to incomplete mesh reconstruction. Consequently, 3D-2D CRF lacks 3D information in these regions and needs to distinguish building instances mainly based on 2D image cues. It is not surprising that the 3D-2D CRF fails at overexposed image regions in this case.

\subsection{Failure Cases}
\begin{figure*}[tb]
\centering
\newcommand{\mywidth}{0.5\textwidth}
\setlength\tabcolsep{0.1em}
\begin{tabular}{cc}

\includegraphics[width=\mywidth]{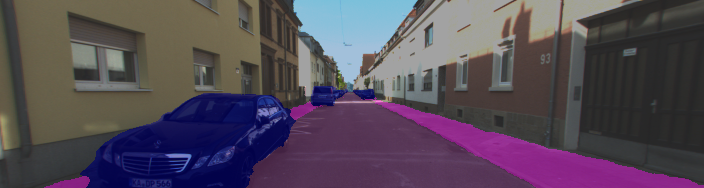} &
\includegraphics[width=\mywidth]{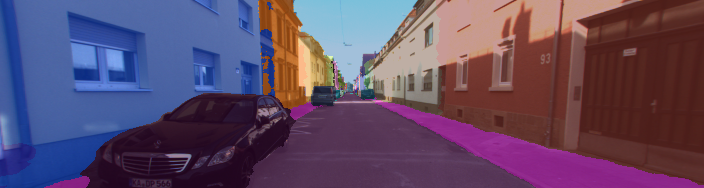}\\
\includegraphics[width=\mywidth]{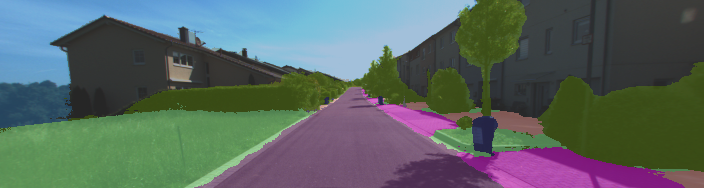} &
\includegraphics[width=\mywidth]{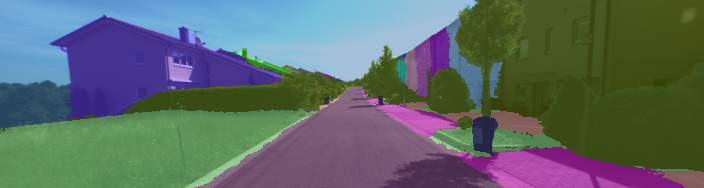}\\
\end{tabular}
\vspace{-0.2cm}
\caption{
\textbf{Failure Cases}. Although our semantic map (left) is correct, the boundary of two adjacent buildings is not well segmented in the panoptic map (right).
}
\label{fig:failure}
\vspace{-0.5cm}
\end{figure*}

Our method leverages a deterministic instance field defined by the 3D bounding primitives to render instance labels. Thus, our method  struggles to recover accurate instance boundaries where two instance bounding primitives overlap in 3D space. This sometimes occurs on the building class where two buildings are spatially connected to each other as shown in \figref{fig:failure}.

\clearpage

\end{document}